%% file: main.tex
\documentclass[journal]{vgtc}
\usepackage{etoolbox}
\patchcmd{\thebibliography}{\clubpenalty4000}{\clubpenalty10000}{}{}
\patchcmd{\thebibliography}{\widowpenalty4000}{\clubpenalty10000}{}{}

\ifpdf
  \pdfoutput=1\relax                   
  \usepackage{graphicx}                
  \DeclareGraphicsExtensions{.pdf,.png,.jpg,.jpeg} 
\else
  \usepackage[dvipdfmx]{graphicx}                
\fi%

\graphicspath{{figures/}{pictures/}{images/}{./}} 

\usepackage{microtype}                 
\PassOptionsToPackage{warn}{textcomp}  
\usepackage{textcomp}                  
\usepackage{times}                     
\usepackage{cite}                      
\usepackage{tabu}                      
\usepackage{booktabs}                  
\usepackage{multirow}
\usepackage{amsmath}
\usepackage{amsfonts}
\usepackage{mathtools}

\usepackage{algorithm}
\usepackage{algpseudocode}
\usepackage{hyperref}
\usepackage[capitalize]{cleveref}
\crefname{figure}{Fig.}{Figs.}

\renewcommand{\ref}{\cref}

\newcommand{\bm}[1]{{\mbox{\boldmath $#1$}}}

\input{macros.tex}

\usepackage{multicol}

\usepackage{xspace}
\newcommand{\ie}{i.e.\@\xspace} 
\newcommand{\cf}{cf.\@\xspace} 

\newcommand{\commentout}[1]{}

\usepackage{color}
\usepackage{savesym} \savesymbol{spacing} 
\usepackage{setspace}
\definecolor{Orange}{rgb}{1,0.5,0}

\definecolor{DarkGreen}{rgb}{0,0.5,0}
\definecolor{Green}{rgb}{0,0.8,0}
\definecolor{Purple}{rgb}{0.8,0,0.8}
\definecolor{Blue}{rgb}{0,0,1.0}
\definecolor{Red}{rgb}{1.0,0.0,0.0}
\definecolor{Brown}{rgb}{0.7,0.4,0.1}

\usepackage{xcolor}

\usepackage{comment}
\usepackage{siunitx} 

\onlineid{0}

\vgtccategory{Research}

\vgtcinsertpkg

\newcommand{\productname}{{PanoTree}}
\title{\productname:~Autonomous Photo-Spot Explorer \\ in Virtual Reality Scenes}

\author{
    Tomohiro Hayase \thanks{e-mail: t.hayase@cluster.mu} \\ %
    \scriptsize Cluster Metaverse Lab
    \and Sacha Braun\thanks{e-mail: s.braun@inria.fr} \\ %
    \scriptsize École Polytechnique
    \and Hikari Yanagawa\thanks{e-mail: h.yanagawa@cluster.mu} \\ %
    \scriptsize Cluster Metaverse Lab
    \and Itsuki Orito\thanks{e-mail: orito.itsuki@gmail.com} \\ %
    \scriptsize Cluster Inc.
    \and Yuichi Hiroi\thanks{e-mail: y.hiroi@cluster.mu} \\ %
    \scriptsize Cluster Metaverse Lab
}

\teaser{
    \includegraphics[width=\linewidth]{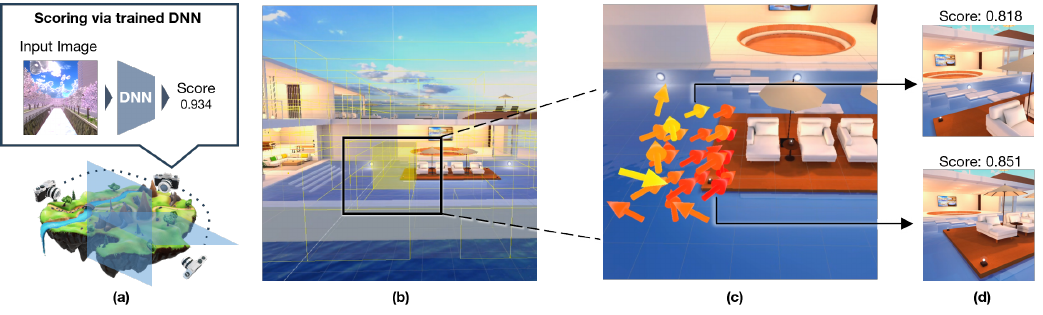}
    \caption{\label{fig:teaser} (a) Our \productname~consists of (top) a DNN-based scoring network that scores images captured in a VR scene, and (bottom) an exploration algorithm that searches for photo spots in the VR scene using HOO, which hierarchically divides the scene and moves the viewpoint to the region where the score of the captured image is higher. (b)Visualization of the divided region of a VR scene by the PanoTree. We visualize the edge of regions in yellow lines, and the yellow box is a selected region with the highest score. (c) Detailed visualization of the selected photo spot. For each coordinate in space, we visualized the direction in which the highest-scoring photo could be taken and the vector field representing that score. The higher the score is, the closer the arrow color is to red. (d) Images taken from the directions suggested by each of the points in (c) and their scores.}
}


\abstract{Social VR platforms enable social, economic, and creative activities by allowing users to create and share their own virtual spaces. In social VR, photography within a VR scene is an important indicator of visitors' activities.
Although automatic identification of photo spots within a VR scene can facilitate the process of creating a VR scene and enhance the visitor experience, there are challenges in quantitatively evaluating photos taken in the VR scene and efficiently exploring the large VR scene.
We propose \productname, an automated photo-spot explorer in VR scenes. To assess the aesthetics of images captured in VR scenes, a deep scoring network is trained on a large dataset of photos collected by a social VR platform to determine whether humans are likely to take similar photos. Furthermore, we propose a Hierarchical Optimistic Optimization (HOO)-based search algorithm to efficiently explore 3D VR spaces with the reward from the scoring network. Our user study shows that the scoring network achieves human-level performance in distinguishing randomly taken images from those taken by humans. In addition, we show applications using the explored photo spots, such as automatic thumbnail generation, support for VR world creation, and visitor flow planning within a VR scene.
}

\CCScatlist{
  \CCScatTwelve{Computing methodologies}{Computer graphics}{Graphics systems and interfaces}{Mixed / augmented reality}
  \CCScatTwelve{Computing methodologies}{Computer graphics}{Graphics systems and interfaces}{Virtual reality}
}

\begin{document}

\input{body}

\bibliographystyle{abbrv}

\raggedbottom
\bibliography{reference}

\newpage
\appendix
\onecolumn

\input{appendix}

\end{document}

%% file: macros.tex
\global\long\def\Vector#1{\mathbf{\MakeLowercase{#1}}}

\global\long\def\Real{\mathbb{R}}
\global\long\def\R{\mathbb{R}}

\global\long\def\RegionAll{\mathcal{X}}
\global\long\def\Region#1{\mathcal{R}_{#1}}

\global\long\def\Depth{d}
\global\long\def\NodeIdx{i}
\global\long\def\Round{n}
\global\long\def\NodeSet{\Depth, \NodeIdx}
\global\long\def\EvalNodeSet{{\Depth}_{\Round}, {\NodeIdx}_{\Round}}
\global\long\def\DepthMax{D_{\EpochNum}}

\global\long\def\Subtree{\mathcal{T}}
\global\long\def\SubtreeIdx#1{\Subtree_{#1}}

\global\long\def\Score{V}
\global\long\def\ScoreN{\Score(\Round)}

\global\long\def\AvgScore#1{\hat{\mu}_{#1}}
\global\long\def\AvgScoreN#1{\AvgScore{#1}(\Round)}
\global\long\def\AvgScoreDef{\AvgScoreN{\NodeSet}}

\global\long\def\Visit#1{T_{#1}}
\global\long\def\VisitN#1{\Visit{#1}(\Round)}
\global\long\def\VisitDef{\VisitN{\NodeSet}}

\global\long\def\UVal#1{U_{#1}}
\global\long\def\UvalN#1{\UVal{#1}(\Round)}
\global\long\def\UvalDef{\UvalN{\NodeSet}}

\global\long\def\BVal#1{B_{#1}}

\global\long\def\DirNum{N_\mathrm{dir}}
\global\long\def\Dir{\boldsymbol{\delta}}
\global\long\def\DirIdx{\Dir_{k}}

\global\long\def\CamNum{N_\mathrm{cam}}
\global\long\def\EpochNum{N}

\global\long\def\DNN{f_\mathrm{S}}
\global\long\def\Renderer{f_\mathrm{R}}

%% file: body.tex
\firstsection{Introduction}
\maketitle
Social VR platforms use virtual reality (VR) technology to facilitate communication between humans through interaction in a 3D virtual environment~\cite{Joshua2019shaping}. These platforms provide a real-time social experience, allowing humans to interact and collaborate with others through avatars. Social VR gained attention, especially during the COVID-19 pandemic~\cite{angeles2022social}, as a new way for people in remote locations to interact in real time. In 2022, more than 171 million humans~\cite{statista2023vr} worldwide access social interaction, economic activity, and creative activity on various social VR platforms such as VRChat~\cite{vrchat2023}, Cluster~\cite{cluster2023}, and Resonite~\cite{resonite2023}.

Social VR allows humans to design and publish hand-made virtual spaces on the platform. These designed virtual spaces are called \emph{worlds}, and a large amount of human-generated content (UGC) is uploaded every day.
Such worlds are expected to be used not only for social functions and games but also for entertainment, such as virtual music concerts~\cite{Hai2018-jl}, education and training~\cite{Makransky2018-nb}, and psychological therapy~\cite{Kaimal2020-je}.

The concept of a VR world gives everyone the ability to design creative spaces beyond the real world.
With free online tutorials, courses, and templates available to the general public, the ability to become a world creator has become a standard on the Internet~\cite{sliwecki2021virtual}.
Like architects in the physical world, VR world creators focus on enhancing the experience for their visitors. Architects design spaces with the visitor's behavior, experience, and satisfaction in mind, often before the space is even visited. Similarly, to help novice creators design VR worlds, social VR platforms should provide tools that take into account expected visitor behavior and experiences to improve human-generated environments.

\begin{figure}[tb]
    \centering
    \includegraphics[ width=\linewidth]{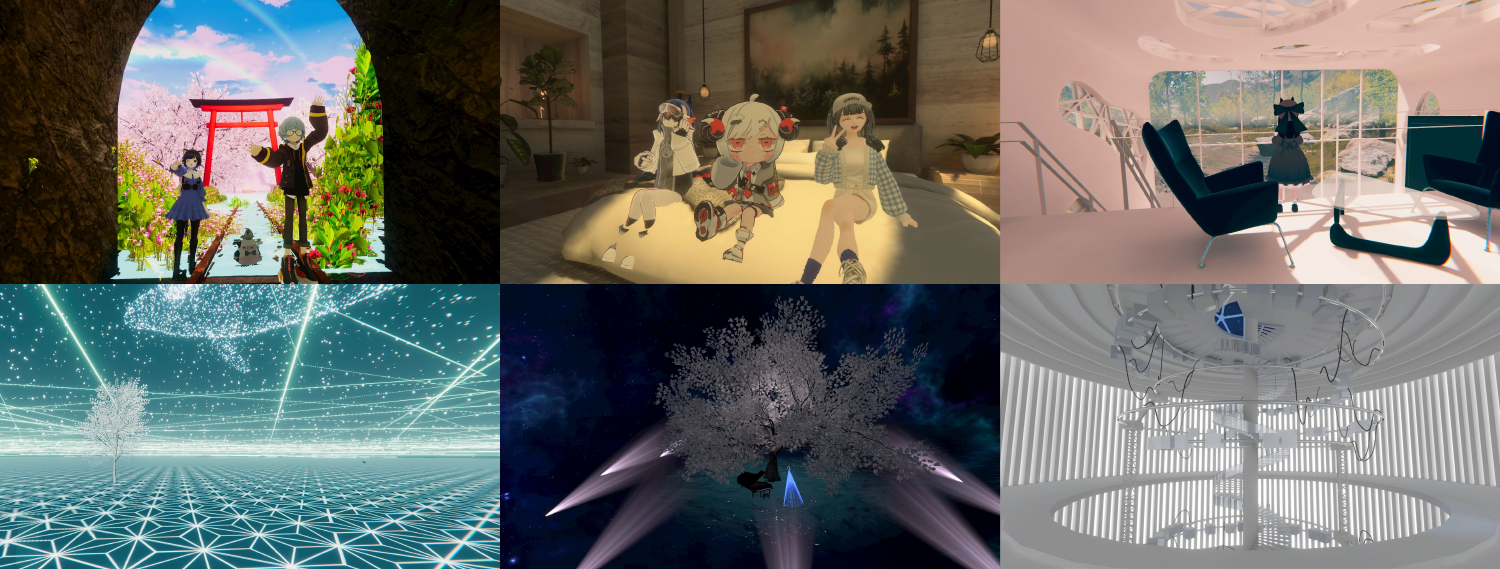}
    \caption{Examples of photos taken by humans in various VR worlds on Cluster, a social VR platform.}
    \label{fig:screenshot-sample}
\end{figure}

We focus on photo spots in  VR worlds as a cue to suggest visitor behavior. As with real-world sightseeing, social VR users find the best parts of the VR world, preserve their memories by taking photos of the landscape, and share their experiences with friends and followers on social networks.
Social VRs have a photo shooting feature to support this motivation, as shown in \autoref{fig:screenshot-sample}. 
In real-world tourism research, crowd-sourced photo data is also gaining attention~\cite{li2018bigdata}. 
Studies have shown that photo spots, identified from the photo data, reveal the most appealing aspects of a location and how visitors interact with it~\cite{Hu2015extracting, Hu2019graphbased, gosal2019social}. Similarly, in VR worlds, analyzing photo spots can significantly aid world designers by providing insights into visitor preferences and guiding better design and marketing strategies.

To realize automatic photo-spot exploration in 3D  scenes of VR worlds, we must address two key challenges: (i) evaluating images captured during exploration to determine their likelihood of being taken by a human and (ii) efficiently searching the entire space of the VR scene for photo spots (\autoref{fig:teaser}, a).
For the former, it is difficult to quantify whether a human is likely to take the input image, especially in a VR scene. 
While deep learning models have successfully assessed the aesthetics of real-world images~\cite{deng2017image, hosu2019effective}, applying similar assessments to VR and 3D computer-generated scenes remains unexplored.
For the latter, searching for photo spots in the world is very computationally intensive.
This process requires varying the position and orientation of the camera - 6 degrees of freedom (DoF) - to evaluate potential photo spots. Performing a grid search over these parameters to identify optimal photo spots involves computational complexity on the order of $O(d^3)$, where $d$ is the number of spatial subdivisions, with more computation required to identify more detailed spots.

In this paper, we propose PanoTree Search, which realizes autonomous photo-spot exploration in VR scenes by solving the aforementioned challenges.
To quantitatively evaluate the aesthetic quality of images captured in VR worlds, we train a deep neural network (DNN) based scoring network (\autoref{sec:photo-evaluator}) that predicts the likelihood of the input image to be captured by humans.
To train this network, we create a large-scale dataset of VR photographs collected from Cluster, a social VR platform in operation (\autoref{sec:method:scoring:data}). 
Subsequently, we propose an algorithm that efficiently explores photo spots within the VR space based on the scores output by the scoring network (\autoref{sec:method:hoo-photospot}). This exploration model approaches the problem of finding 3D photo spots as a multi-armed bandit problem in continuous space and efficiently explores photo spots across the entire scene using hierarchical search. Our PanoTree algorithm is implemented in a platform-agnostic manner (\autoref{sec:impl}), making it applicable to any 3D scene. We evaluate the scoring network and the explored photo-spots  (\autoref{sec:eval}), demonstrating the potential for practical applications (\autoref{sec:appliaction}).

Our primary contributions are that we:
\begin{itemize}
    \vspace{-1.5mm}
    \item Introduce \productname, an automated photo-spot explorer in VR space, to enhance the experience of user-generated VR content on the VR platform.
    \vspace{-1.5mm} 
    \item Develop a scoring network trained on a wild large photo dataset collected from social VR platforms in operation.
    \vspace{-1.5mm}
    \item Develop an efficient spatial hierarchical search algorithm for photo-spot exploration with implicit supervision.
    \vspace{-1.5mm}
    \item Implement and evaluate a prototype for our PanoTree search, discuss exploration quality, and provide potential applications.
\end{itemize}

\section{Related Work}

\subsection{Black-box Optimization}
We treat the evaluation function of photography spots within a space as a black-box function with respect to camera coordinates for black-box optimization. Black-box optimization can be approached by various methods \cite{powell1964efficient, nelder1965simplex, brochu2010tutorial, bubeck2011xarmed, eiben2015introduction, huang2023global}. Among these, we use hierarchical optimistic optimization (HOO)~\cite{bubeck2011xarmed}, which is well-suited for deep learning models and Euclidean spaces.

HOO is a method that applies the multi-armed bandit problem to general measurable spaces, such as Euclidean spaces, to find the maximum value of a black-box function over continuous spaces. The multi-armed bandit problem is a fundamental problem setting in reinforcement learning for balancing exploration and exploitation. The upper confidence bound (UCB) search is a basic solution for the bandit problem, and the upper confidence trees (UCT) search is its adaptation to tree structures. Unlike standard UCT, HOO tightens the upper confidence bound by making child nodes explicit subregions of the parent node's region. However, directly using HOO can result in a high computational cost of $O(n^2)$, where $n$ is the number of exploration iterations. Our algorithm is based on the truncated HOO proposed by \cite{bubeck2011xarmed}, which is a version of HOO with a reduced computational cost of $O(n\log n)$.

Furthermore, truncated HOO is a deterministic method. However, deterministic division might miss small but important spots in the search for photography spots, requiring extended search times. To address this issue, we propose a probabilistic division called softmax policy, which helps to avoid this problem.

\subsection{Semantic Understanding and Exploration in 3D Space}
Understanding 3D scenes is a problem that has traditionally been addressed~\cite{roberts1963machine, yakimovsky1973semantics}. In particular, recent developments in DNN have made it possible to handle large-scale media data such as 3D scenes. 3D semantic segmentation~\cite{garcia2018survey, minaee2021image} is an application of DNN to understanding 3D scenes, which detects object labels in a real space along with their positions. It is intended to distinguish objects in space and differs from our work, which predicts spots in 3D space to perform a specific task.

\subsection{Support for VR Scene Creation}
VR scene creation requires both artistic and technical skills and involves managing a large number of variables. For this reason, numerous types of research have been proposed to support the creation of VR or 3D scenes~\cite{Oliveira2023Virtual}. Initially, automated VR and 3D scene creation relied on guideline-based layout designs~\cite{Merrell2011furniture, Yu2011makeithome}. This evolved into data-driven methods that use large 3D datasets~\cite{song2017suncg} to refine object placement~\cite{Wang2018indoor} and combine existing 3D assets in VR worlds~\cite{ma2018language}.

Recent developments in large language models (LLMs) and multimodal machine learning allow for more intricate modifications through natural language prompts, such as 3D object generation~\cite{xu2023dream3d, poole2023dreamfusion, lin2023magic3d, tang2023dreamgaussian}. Methods have also been proposed to ensemble these models to create designs for entire VR worlds~\cite{yin2024text2vrscene} and to allow interactive editing of VR worlds~\cite{sun20233dgpt}. Furthermore, beyond 3D objects, methods have been proposed to create interactive elements in VR spaces by generating scripts from natural language using LLMs~\cite{roberts2022steps, giunchi2024dreamcode}. More recently, LLMR proposed a generic framework for applying LLMs to generate interactive 3D experiences by ensembling several specialized GPTs, including a builder, a scene analyzer, a skills library, and an inspector~\cite{delatorre2024llmr}.

Our method differs from the LLM-based approach in that we use a large amount of human-generated non-verbal behavioral data to support the creation of VR worlds. To our knowledge, this is the first study to use accumulated data in a social VR platform to support the creation of VR worlds. In the future, we expect to integrate non-verbal behavioral data collected in social VR platforms with LLMs containing language-based general knowledge to support higher-order creation in VR.

\section{Dataset}\label{sec:method:scoring:data}

\begin{figure}[t]
    \centering
    \includegraphics[width=\linewidth]{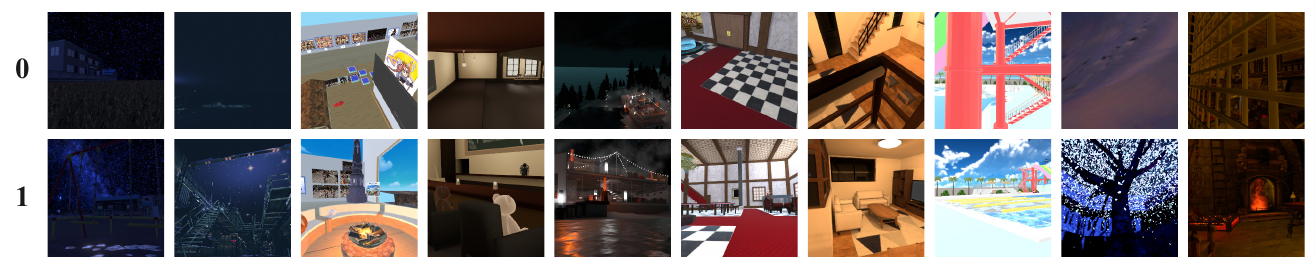}

    \includegraphics[width=\linewidth]{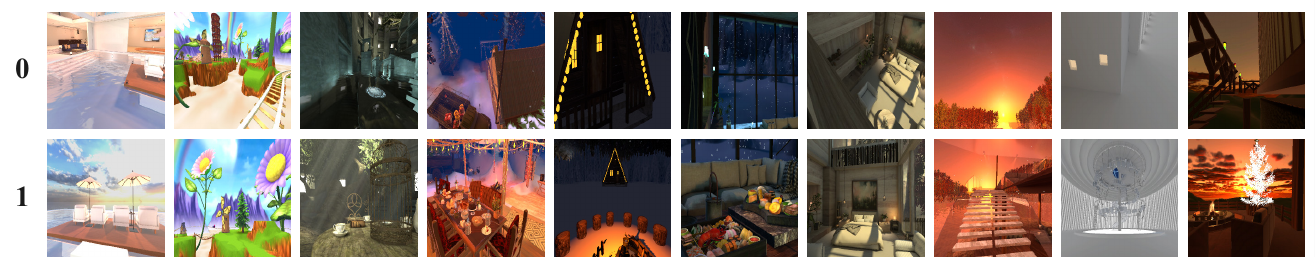}
    \caption{Pairs of negative samples (0)  and positive samples (1)  from ten test scenes in WI-89 (top) and WI-245 (bottom).}
    \label{fig:test_scenes}
\end{figure}

We provide an overview of the dataset of VR scenes and images used for our exploration before delving into the details of the PanoTree.
The dataset consists of two main components: a collection of images for training the scoring function of photography spots and a VR scene designed for evaluating exploration algorithms.

In VR scenes, photos taken by humans often include avatars that can affect the evaluation of photography spots. When creating VR worlds and identifying photography spots, it is crucial to evaluate these spots without the influence of avatars. Therefore, images used for training the scoring function of photography spots need to exclude avatars.

To achieve this, we utilize the accumulated photography metadata on the VR platform Cluster to recreate the shooting conditions and retake the photos without the avatars. Similar to how people choose backgrounds in real-world photography, we hypothesize that humans also select backgrounds in VR scenes. Hence, we believe that even when avatars are excluded, the recaptured images will still contain relevant information about the photography spots.

\subsection{Metadata Collection and VR Scene Selection}\label{ssec:metadata}
To build our dataset, we first collected photography metadata from the VR platform Cluster\cite{cluster2023}. This metadata includes the camera's position and orientation, field of view (FoV), and the VR scene used for taking each photo. When equipped with VR trackers, users can directly manipulate the arm of the VR avatar while holding a camera to take pictures. Additionally, the camera can be placed at a distance from the avatar's position, allowing for high flexibility in height.

For the selection of VR scenes, we gathered the top-k VR worlds based on the number of photos taken during specific periods, excluding VR worlds with scene transitions. Therefore, each VR world in the collection consists of a single 3D scene. Over a period of four months (resp.\,one year), we identified the top-100 (resp.\,top-300) VR worlds, resulting in 89 (resp.\,245) distinct worlds. We denote these datasets as WI-89 (resp.\,WI-245), which include the VR worlds and corresponding images collected by the methods described in \autoref{ssec:dataset_creation}.  The larger and more diverse dataset WI-245 is instrumental in our study's objectives.

\subsection{Dataset Creation}\label{ssec:dataset_creation}
We collected all physical collision regions associated with objects for each VR scene. Then, we set $\RegionAll$ as the smallest bounding box encompassing all collision regions for each VR scene. We use $\RegionAll$ as the space for exploration algorithms.

We captured negative images labeled as 0 by setting a random camera position and orientation according to the region collisions of each VR scene contained in the datasets. We captured positive images labeled as 1 using metadata from human photo-shooting in each VR scene contained in the datasets. In the capturing process, we excluded avatars and only included the objects and backgrounds in the scenes. \autoref{fig:test_scenes} shows a selected pair of positive and negative images from ten scenes in WI-89 and WI-245.

\section{PanoTree Search}
The PanoTree Search consists of a scoring network and an exploration algorithm.

Our explorer searches for photo spots in the 3D CG space based on the score of the trained scoring network.
To make the search efficient, we use Hierarchical Optimistic Optimization (HOO)~\cite{bubeck2011xarmed} to find the point that maximizes the score. 
The HOO is a black-box optimization strategy that extends bandit algorithms and is effective for problems with Euclidean search spaces and unknown or deep score functions. The HOO is also effective for our problem since our photo-spot exploration consists of a renderer without a differential function and optimizing a DNN-based complex score function for camera poses.

We introduce spatial division of the VR space and region evaluation in the HOO to apply the HOO to photo-spot exploration.
First, we describe the general HOO strategy (\autoref{sec:method:hoo}) and then our HOO strategy for photo-spot exploration (\autoref{sec:method:hoo-photospot}).

\subsection{Photo Scoring Network in VR Scene}\label{sec:photo-evaluator}

\begin{figure}[tb]
    \centering
    \includegraphics[width=\linewidth]{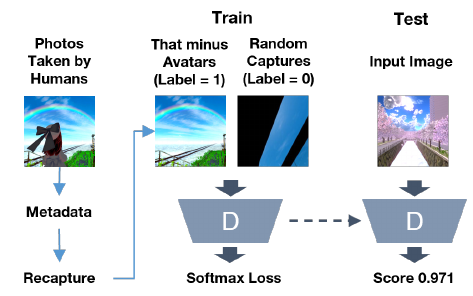}
    
    \caption{Scoring network of the photo in VR space. The network learns whether the input image is human-captured or randomly captured and features the image likely to be captured by a human. The input images are labeled with  1 for human-captured and 0 for randomly captured. During the evaluation, for each input image, the network outputs a score indicating whether the image is likely to have been captured by a human. 
    }
    \label{fig:scoring-network-overview}
\end{figure}

Next, we describe the network training process to score the input images using the dataset detailed in \autoref{sec:method:scoring:data}.
\autoref{fig:test_scenes} shows scenes for evaluation of the testing accuracy of scoring networks and the exploration algorithm.
%
\autoref{fig:scoring-network-overview} provides an overview of our scoring network. Using our dataset, we train a scoring network $\DNN:\Real^{HWC} \to [0, 1]$ that outputs the probability that a human would capture an input image $X \in\Real^{HWC}$, where $H$ and $W$ represent the height and width of the image, respectively, and $C$ denotes the number of channels.

The scoring network $\DNN$ is trained as a binary classifier using cross-entropy loss, with random photos labeled as 0 and photos taken by humans labeled as 1. When an unknown photo is input, $\DNN$ outputs a value in the range [0, 1]. The closer this value is to 1, the more likely the photo is to resemble those taken by humans, which is considered the score of the input photo. A detailed implementation of the network will be described in \autoref{sec:impl:scoring}.

\subsection{{Truncated HOO Strategy}}\label{sec:method:hoo}
\begin{figure*}[tb]
    \centering
    \includegraphics[width=0.625\linewidth]{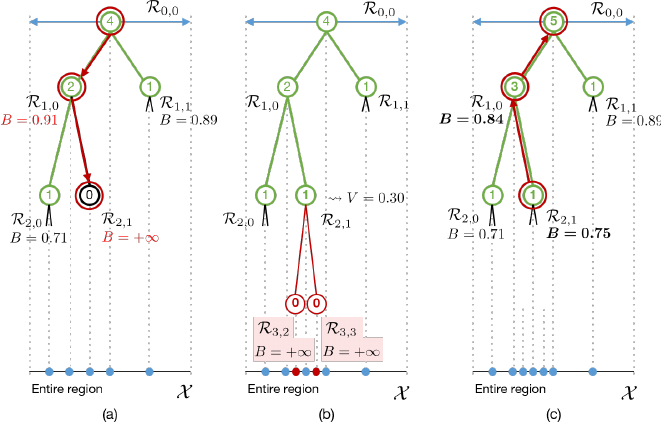}
    \includegraphics[width=0.365\linewidth]{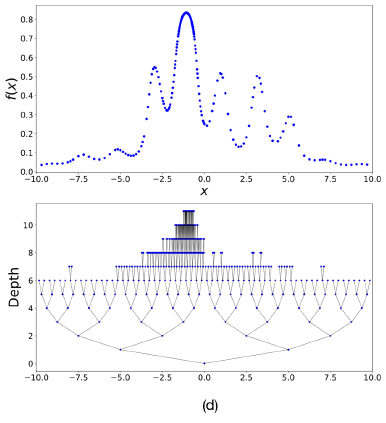}
    \caption{Overview of the truncated HOO.  The number on each node indicates the number of visits. (a) At the beginning of each iteration, starting from the root node, the explorer traverses the tree by selecting the child with the highest B-value and arrives at the leaf (red arrows). (b) After including the arriving leaf nodes in the tree, the space corresponding to the arriving leaf is split, and two child nodes (red nodes) are added. (c) Update of the $B$-value. Starting from the leaf node marked by a red circle, the B-values and number of visits are updated up to the root node (red arrows). Once the update is complete, the explorer moves on to the next iteration. In this example, the explorer will arrive at node (1, 1) in the next iteration. (d) Results of the truncated HOO with $f:\Real\supset\RegionAll=[-10,10]\rightarrow\Real$ as an example (top) and its tree (bottom).}
    \label{fig:hoo-tree-schematic}
\end{figure*}

\begin{algorithm}[t]
\caption{Truncated HOO Strategy}
\label{alg:hoo}
\textbf{Auxiliary functions:} \\
\hspace*{\algorithmicindent}\Call{Dividespace}{$\Region{}$}: outputs $\Region{}$ in two parts. (see \autoref{sec:method:hoo:division}) \\
\hspace*{\algorithmicindent}\Call{Evalspace}{$\Region{}$}: outputs a reward of $\Region{}$. (see \autoref{sec:method:hoo:reward})\\
\textbf{Initialization:} \\
\hspace*{\algorithmicindent}$\Region{0,0} = \RegionAll$ and $\Subtree=\{(0,0)\}$. \\
\hspace*{\algorithmicindent}$\Region{1, 0}, \Region{1,1} = $ \Call{Dividespace}{$\Region{0, 0}$}.\\
\hspace*{\algorithmicindent}$\BVal{1,0}{}=\BVal{1,1}{}=+\infty$.\\
\hspace*{\algorithmicindent}$T_{1,0}=T_{1,1}=0$.
\begin{algorithmic}[1]
\Procedure{HOO}{$c, \nu_1, \rho, \EpochNum $}
\For{$n=1,2,\cdots, \EpochNum$} 
    \State $(\NodeSet)\leftarrow(0,0)$  \label{alg:l:s-search}
    \While{$(d,i) \in \Subtree $ and $\Depth< \DepthMax$} \Comment{Selection}
        \State $T_{\Depth,\NodeIdx} += 1$
        \If{$\BVal{\Depth+1, 2\NodeIdx}>\BVal{d+1, 2i+1}$} \label{alg:l:s-node-select}
        \State $(\NodeSet)\leftarrow(\Depth +1, 2 \NodeIdx)$
        \ElsIf{$\BVal{\Depth +1, 2\NodeIdx}<\BVal{\Depth+1, 2\NodeIdx+1}$}
        \State $(\NodeSet)\leftarrow(\Depth+1, 2\NodeIdx+1)$ \label{alg:l:e-node-select}
        \Else \label{alg:l:s-random-select}
        \State Sample $z$ from $\mathrm{Uniform}(\{0,1\})$ 
        \State $(\NodeSet)\leftarrow(\Depth +1, 2\NodeIdx+z)$
        \EndIf \label{alg:l:e-random-select}
    \EndWhile \label{alg:l:e-search}
    \State $T_{d,i}+=1$.
    \State $(\EvalNodeSet) \leftarrow(\NodeSet)$ 
    \State $\Subtree\leftarrow\Subtree\cup\{(\Depth, \NodeIdx)\}$ \label{alg:l:s-divide-space}  \Comment{Expansion}

    \State $\Region{\Depth+1, 2\NodeIdx}, \Region{\Depth +1,2\NodeIdx+1} = $ \Call{DivideSpace}{$\Region{\Depth,\NodeIdx}$}. \label{alg:l:s-divide-region}
    \State $\BVal{\Depth+1, 2\NodeIdx} = \BVal{\Depth +1, 2 \NodeIdx+1} = +\infty$ \label{alg:l:new-leaf-infty}.
    \State $T_{\Depth +1, 2\NodeIdx}=T_{\Depth+1, 2\NodeIdx+1}=0$.\label{alg:l:e-divide-region}

    \State $\Score \leftarrow$\Call{Evalspace}{$\Region{\EvalNodeSet}$} 
    \label{alg:l:s-eval-region} \Comment{Simulation}

    \While{ $d >0$} \Comment{Backup}
        \State $\AvgScore{\NodeSet}\leftarrow(1-1/\Visit{\NodeSet})\AvgScore{\NodeSet}+\Score/\Visit{\NodeSet}$ \label{alg:l:update-avg}
        \State $\UVal{\NodeSet}\leftarrow\hat{\mu}_{\NodeSet}+c\sqrt{2\log{\EpochNum}/\Visit{\NodeSet}}+\nu_1 \rho^\Depth$\label{alg:l:update-U}

        \State $\BVal{\NodeSet}\leftarrow\min \left(\UVal{\NodeSet}, \max\left(\BVal{\Depth+1,2\NodeIdx}, \BVal{\Depth +1,2\NodeIdx+1}\right)\right)$\label{alg:l:update-B}
        \State \Comment{Tighter Upper Bound}
        \State $(\Depth,\NodeIdx) \leftarrow (\Depth-1, [\NodeIdx/2])$ \Comment{Go to the parent node}
    \EndWhile \label{alg:l:backprop}
    \EndFor
\State \Return $\Subtree$
\EndProcedure
\end{algorithmic}
\end{algorithm}

\autoref{fig:hoo-tree-schematic}  shows an overview of the truncated HOO, and Algorithm~\ref{alg:hoo} provides the pseudo-code of the truncated HOO. The truncated HOO performs a binary tree search by hierarchically dividing the search space, balancing the exploitation and exploration of sub-regions. The region associated with each node is evaluated using the Upper Confidence Bound (UCB)~\cite{Peter2003ucb} for Euclid spaces, which combines the score obtained from the scoring function and the uncertainty due to the small number of searches in the region. In this section, after introducing the mathematical notation (see \autoref{tab:notaions} for a summary of notations), we describe UCB-based search, region division, and evaluation.

\subsubsection{Notations}

We associate an entire search space box with the tree's root node, and each node in the tree is associated with a measurable subset of the entire space. We call this subset a region.  The nodes in this tree are indexed by pairs of integers $(\NodeSet)$; $\Depth \geq 0$ is the depth of the node, and $i=0, \cdots,2^{\Depth}-1$ is the index of the node. 

Let $\RegionAll$ be the entire search space and $\Region{\Depth,i}$ be the region associated with node $(\Depth,i)$. Since the root node $(0,0)$ corresponds to $\RegionAll$, we have
\begin{align}\label{align:region-division}
\Region{0,0} &= \RegionAll, \nonumber \\
\Region{\Depth,\NodeIdx} &= \Region{\Depth+1, 2\NodeIdx}\cup\Region{\Depth +1,2\NodeIdx+1}.
\end{align}
Now, for any $\Depth$, we assume that $\Region{\Depth, \NodeIdx} (\NodeIdx =0, 1, \dots, 2^{d}-1 )$ are mutually disjoint sets and $\cup_{\NodeIdx=0}^{2^\Depth -1 } \Region{\Depth, \NodeIdx} = \RegionAll$.

Let us formally introduce time-indexed notations for quantities in the algorithm.
The indexing by $n$ is used to indicate the value taken at iteration $n$.
For each $n$, we have a tree $\SubtreeIdx{n}$. At the initial state, we set $\SubtreeIdx{0}=\{(0,0)\}$. For node selections, we assume that each tree $\SubtreeIdx{n}$ is equipped with values $B_{\NodeSet}(n) \in \R\cup\{+\infty\}$, which is an estimated upper bound in the mean scoring function over the region $\Region{\NodeSet}$.

\subsubsection{Selection and Expansion}
At the beginning of each exploration iteration, the truncated HOO selects the most promising action for the explorer (\cf lines \ref{alg:l:s-search}--\ref{alg:l:e-search} of Algorithm~\ref{alg:hoo}). Starting at the root node, the explorer traverses the tree until it reaches an undivided region $\Region{d_n, i_n}$, always selecting the child node with the highest $B$-value (\autoref{fig:hoo-tree-schematic} (a) and lines \ref{alg:l:s-node-select}--\ref{alg:l:e-node-select}).
If two child nodes have the same $B$-value, the explorer chooses a child uniformly (lines \ref{alg:l:s-random-select}--\ref{alg:l:e-random-select}). 
Then we add the node to the tree as follows:
\begin{align}
    \SubtreeIdx{n}=\SubtreeIdx{n-1}\cup\{(\EvalNodeSet)\},\,  n\geq 1.
\end{align}

After the leaf node $(\EvalNodeSet)$ is included in the post-visit tree $\SubtreeIdx{n}$, we divide the region corresponding to the node (lines \ref{alg:l:s-divide-region}--\ref{alg:l:e-divide-region}, see \autoref{sec:method:hoo:division} for our division in detail).  Note that we do not add the child nodes into the tree at this time.

\subsubsection{Upper Confidence Bound}
In this section,  we define the $B$-value at each node, which is an estimated upper bound of mean rewards at the corresponding region.
To define the $B$-values, we need to introduce the number of played descendants of the node $(\NodeSet)$, denoted by $T_{\NodeSet}(n)$.  It holds that a node in $\SubtreeIdx{n}$ is a leaf if and only if $T_{\NodeSet}(n)=1$.

  Next, we introduce the average of the reward of descendants received during exploration as follows:
\begin{align}
    \AvgScoreDef =  \frac{1}{\VisitDef}  \sum_{ \substack{ m =0,1,\dots,n, \\ \Region{d_m^*, i_m^*} \subset \Region{d,i} }} V(m),
\end{align}
where $\ScoreN$ is the reward in iteration $\Round$. In particular, it corresponds to the output of the scoring function (\autoref{sec:method:hoo:reward} in detail.).


For any $(\NodeSet) \in \SubtreeIdx{n}$, we define
\begin{align}\label{align:uvaldef}
    \UvalDef =   \AvgScoreDef + c \sqrt{ 2 \log \EpochNum \over \VisitDef} + \nu_1 \rho^d,
\end{align}
where $c >0,  \nu_1 >0 ,   0< \rho \leq 1$ are hyperparameters. The last term of \eqref{align:uvaldef}, $\nu_1 \rho^\Depth$, is the regularization term that characterizes the smoothness of the target function.

Last, we define inductively the $B$-value, which is a tighter estimated upper bound in the mean target function at the node $(\NodeSet)$ than the $U$-value.  For each leaf node, we set $B_{\NodeSet}(n)= U_{\NodeSet}(n)$.  For each non-leaf node, we set 
\begin{align}\label{align:Bval-def}
    B_{\NodeSet}(n) = \min \{ U_{\NodeSet}(n), \max \{ B_{d+1, 2i}(n), B_{d+1, 2i+1}(n)  \} \}.
\end{align}
The definition is based on two observations:  the nodes represent the region division  \eqref{align:region-division}, and the $U$-value is an estimate of the upper bound.

\subsubsection{Backup}
Once the explorer reaches a leaf node $(\EvalNodeSet)$, the region $\Region{\EvalNodeSet}$ is evaluated (see \autoref{sec:method:hoo:reward} in detail) and the reward $\ScoreN$ is obtained (\autoref{alg:l:s-eval-region}). 
 By convention, we set $T_{d_n+1, 2i_n}(n)=T_{d_n+1, 2i_n+1}(n)=0$ and $B_{d_n+1, 2i_n}(n)=B_{d_n+1, 2i_n+1}(n)=+\infty$.
Consider a node $(\NodeSet)$ on the traversed path at the iteration $n$. 
The number of visits and the averaged reward are updated  as follows (\autoref{alg:l:update-avg}):
\begin{align}
\VisitDef&=\Visit{\NodeSet}(\Round-1)+1, \nonumber \\
\AvgScoreDef&=\cfrac{\Visit{\NodeSet}(n-1)\AvgScore{\NodeSet}(n-1)+\ScoreN}{\VisitN{\NodeSet}}. \nonumber
\end{align}
We update the $U$-value according to \autoref{align:uvaldef}.
Finally, $B$-values are backward updated  from the $(d_n, i_n)$ to the root node by \eqref{align:Bval-def} as \autoref{fig:hoo-tree-schematic} (c) and \cref{alg:hoo}~(line~\autoref{alg:l:update-B}).

For the nodes out of the traversed path, we set the values $T, \hat{\mu}, U$, and $B$  the same as those at the previous iteration $n-1$  since  $U$ only depends on the last visit.

\subsubsection{Depth Limitation}
The depth limitation $D_{\EpochNum}$ of the tree, as given in\cite{bubeck2011xarmed}, is as follows:
\begin{align}\label{align:depth-limit}
    D_{\EpochNum} = \lceil {  \log \EpochNum + \log v_1 \over  - \log \rho  } \rceil,
\end{align}
where $\lceil \cdot  \rceil$ is the ceiling function.
 
\subsubsection{One-Dimensional Example}
\autoref{fig:hoo-tree-schematic} (d) shows an example of HOO exploration in a 1D space $[-10,10]$. The regions near the maximum were explored more frequently, and the nodes there were deeper than the others.

\subsection{Explorer Part of PanoTree}\label{sec:method:hoo-photospot}
In this section, we introduce the explorer of the PanoTree based on the truncated HOO.
We introduce spatial division policies and a region evaluation method to the truncated HOO for photo-shot exploration.

\subsubsection{Search Space}\label{sec:method:hoo:search_space}
The camera parameter to be optimized is the 6 degrees of freedom (6DoF). Since position and orientation have completely different properties, we will apply the HOO search only to the position. Therefore, we define the overall search space $\RegionAll$ as a cuboid within the 3D space determined by the bounding box given by \autoref{ssec:dataset_creation}. Consequently, the divided regions are also 3D cuboids. The camera is placed at the center of each region. We denote the center of $\Region{\NodeSet}$ by $\Vector{c}_{\NodeSet}$. \autoref{fig:hoo-photography}~(a) shows an overview of the region division in the PanoTree.

\subsubsection{Spacial Division Policy}\label{sec:method:hoo:division}
Consider a region  $\Region{\NodeSet}$ to be divided. 
We introduce spatial division policies based on the length of the region's edges to balance the axes.
Let us denote the length vector of the region's edges by $L=(L_x, L_y, L_z)$. 

\textbf{Argmax Policy:}  A division policy is to divide the edge of the region which has maximal length. We call this the \emph{argmax policy}.

\textbf{Softmax Policy:}  To avoid a strong dependence of the exploration on slight differences between edge lengths, we introduce the \emph{softmax policy} defined as follows.  Consider a probability distribution $\pi$ on $(x,y,z)$ obtained by the softmax function below:
\begin{align}
    \pi_i =  \cfrac{\exp(L_i/||L||)}{\sum_{j=x,y,z} \exp(L_j/||L||)}, \quad i=x,y,z,
\end{align}
where $||L||$ is the Euclidean norm. We sample the index of the edge to be divided from the distribution $\pi$.  

We set the softmax policy as the default policy of the PanoTree's explorer.

\subsubsection{Region Evaluation}\label{sec:method:hoo:reward}
\begin{figure}[tb]
    \centering
    \includegraphics[width=\linewidth]{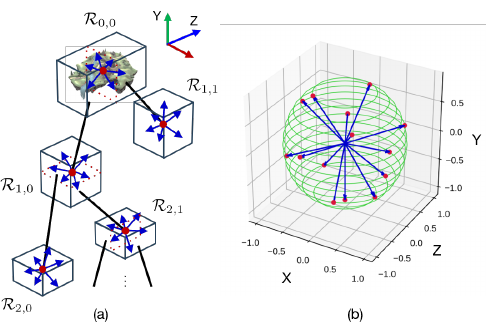}
    \caption{(a) 3D Spatial division and tree structure of the PanoTree. Each scene is defined as a cuboid and is divided so that the longest edge is the most likely to be divided. (b) Sampled directional vectors $\DirIdx$ (blue arrows) in $\DirNum=15$. Each $\DirIdx$ is a unit vector whose destination (red) is distributed over the unit sphere (green).}
    \label{fig:hoo-photography}
\end{figure}

Consider the node $(\EvalNodeSet)$ to be evaluated in the iteration $\Round$. We take multiple directions from the camera center to evaluate the region and aggregate the information using the scoring function.
In each region, the $\DirNum$ direction vectors $\DirIdx=[x_k, y_k, z_k]^\top  (k=0,1,\cdots\DirNum-1)$ are defined as follows:
\begin{align}
\label{eq:hoo-direction-sample}
x_k=r\cos(\theta),\;\; y_k = 1-\cfrac{2k}{\DirNum-1},\;\; z_k=r\sin(\theta), \\
\mathrm{where}\quad r=\sqrt{1-y_k^2}, \quad \theta = (1+\sqrt{5})\pi k. \nonumber
\end{align}
\autoref{fig:hoo-photography} (b) shows $\DirIdx$ sampled by (\ref{eq:hoo-direction-sample}), which confirms that the sampled direction vectors are sampled almost uniformly around the point.

Let $\Renderer(\Vector{c}_{\EvalNodeSet}, \DirIdx)$ be the rendered image of the scene from the center point $\Vector{c}_{\EvalNodeSet}$ and the direction vector $\DirIdx$. We set the reward $\ScoreN$ as follows:
\begin{align}
\label{eq:reward-by-sampled-dirs}
\ScoreN = \max_{k=0, \dots, \DirNum-1} \DNN\left(\Renderer\left(\Vector{c}_{\EvalNodeSet}, \DirIdx \right) \right).
\end{align}

\subsubsection{Discussion}\label{sscec:discussion}
Note that, since most VR platforms use game engines such as Unity, the renderer is generally not differentiable. Therefore, the direction vector of the camera cannot be continuously optimized to maximize the reward as in differentiable rendering, which limits the resolution of the optimized direction vector in the current system (\autoref{sec:discuss:nerf}).

The truncated HOO is a computationally efficient version of the vanilla HOO, referred to simply as the HOO in \cite{bubeck2011xarmed}. The vanilla HOO uses the $U$-value defined by
\begin{align}
    \UvalDef = 
        \AvgScoreDef + c \sqrt{ \cfrac{ 2\log{n} }{ \VisitDef } }+\nu_1 \rho^\Depth.
    \label{eq:update-u-vanilla}
\end{align}
In that case, the term $\sqrt{ 2\log{n} / \VisitDef  }$ depends not only on the last visit but also on $n$.  Therefore, as mentioned in \cite{bubeck2011xarmed}, the backup of $B$-values in the vanilla algorithm requires the $U$-value of nodes not on the traversal path. This results in a computational complexity of $O(n^2)$. 
Instead, in the truncated HOO, the $U$-value of nodes out of the traversal path is determined by $N$ instead of $n$ for any $(\NodeSet) \in \SubtreeIdx{n}$. This adjustment ensures backup along a single path and that the update of $B$-values on the traversal path can be accomplished in $O(n\log n)$.

The depth limit \eqref{align:depth-limit} is due to suppressing deeper nodes than $O(\log n)$. Since the suitable depth limit can depend on the scene, we set $D_T = \infty$ as the default for the PanoTree's explorer.

\section{Implementation}\label{sec:impl}
In this section, we describe the detailed implementation of the dataset and the \productname.

\subsection{Dataset}
During random capture, the range of the camera position is limited to within bounding boxes where objects are located, excluding positions more than 25 meters above the ground. The FoV of the virtual camera is set at 60$^{\circ}$.  
As shown in \autoref{tab:dataset}, we collected reconstruction of $0.21 \times 10^6$ (resp.\,$0.83 \times 10^6$) images from metadata of human photo-shooting for WI-89 (resp.\,WI-245). 
When we recapture positive images from metadata, we set the FoV as 60$^{\circ}$ to make them consistent with negative samples. Additionally, we exclude positive images with a standard deviation of less than 0.05 to remove images of only floors, ceilings, or skyboxes.
In total, the dataset consisted of $0.40 \times 10^6$ (resp.\,$1.65 \times 10^6$) images in two categories.

We separate the scenes into training and test sets.  \autoref{tab:dataset_detail} shows the detailed statistics regarding this separation. The hyperparameters of the scoring function are determined by further splitting the training scenes. 
Additional human annotation was used to remove false-positive and false-negative images from the test image dataset to ensure data quality.
For the scenes used to test the exploration algorithm, the same scenes used to create the test images are utilized to prevent leakage.

\begin{table}[t]
    \centering
    \caption{Dataset statistics of VR spaces about the number of scenes, the number of images, the number of humans' photo-shooting metadata, and the average volume of the search space $\RegionAll$.}
    \begin{tabular}{c|ccccc}
    \toprule
    Datasets &  Scenes & Images & Metadata & Avg.\,Vol.\,(\si{m^3})\\
    \midrule
       WI-89  &  89  & $0.40 \times 10^6$    &  $0.21 \times 10^6$   & $4.6\times 10^8$  \\
       WI-245 &  245 & $1.65\times 10^6$  &  $0.83 \times 10^6$ &    $7.4\times 10^8$ 
    \end{tabular}
    \label{tab:dataset}
\end{table}

\subsection{Scoring Network}\label{sec:impl:scoring}
As a scoring network, we compared two models: Vision Transformer (ViT-B/16)~\cite{dosovitskiy2021an} and MLP-Mixer (Mixer-B/16)~\cite{tolstikhin2021mlp}. Both models were pre-trained by extensive training on ImageNet-21k and fine-tuning on ImageNet-1k. To train for binary classification, we replaced the final linear layer of each model with a 2D output. We then fine-tuned each model on our dataset for five epochs with batch size 256. We used AdamW as an optimizer and set its learning rate to $10^{-5}$ with cosine annealing scheduling. The learning rate was warmed up during the first epoch with a starting learning rate of $10^{-6}$. Dropout and label smoothing rates were set at 0.5 and 0.1, respectively. The settings of hyperparameters are summarized in \autoref{tab:aug_hps}.
We implemented the model using PyTorch as the framework.

To determine the appropriate network for this problem, the same data set was trained on ViT-B/16 and Mixer-B/16. The results were $83.95 \pm 0.38 \%$ for ViT-B/16 and $82.14\pm 0.20 \%$ for Mixer-B/16. 
Each model was trained with eight different random seeds, and thus, the experiment results summarize eight different models.
Based on these results, ViT-B/16 is used as the scoring network in subsequent experiments.

\subsection{Exploration}

\begin{figure}[t]
    \centering
    \includegraphics[width=0.488\linewidth]{fig/rendering_000000.pdf}
    \includegraphics[width=0.5\linewidth]{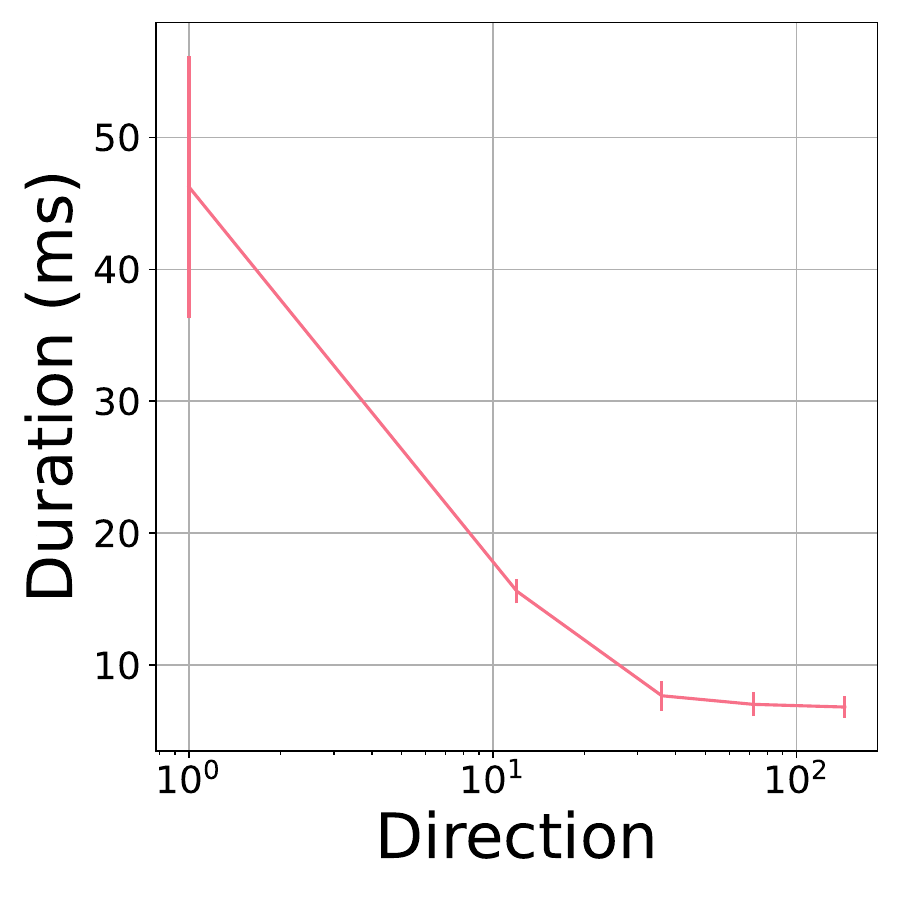}
    \caption{Parallel rendering and evaluation via server-client communication with $\CamNum=36$.  (Left) Images captured by multiple cameras are tiled and combined into a single image. This image is transferred to the Python client as a byte sequence. (Right) The average duration per image and iteration  The results are averaged over ten scenes and three trials with different random seeds. 
    }
    \label{fig:parallel}
\end{figure}

For a reduction in evaluation time of the space, we use parallel rendering. We use a rendering server to render and capture images in the VR scene, which are requested by the Python client to be passed to the scoring network. We denote by $\CamNum$ the maximal number of images per rendering loop to be passed.

In the implementation, the explorer on the Python client requests the capture of images for $\DirNum$ shooting directions in ((\ref{eq:reward-by-sampled-dirs}), and the results are batch-processed on the Python client to speed up the exploration. The state of the tree, including space information and camera parameters, is managed by the Python server.

As an initialization, $\CamNum$ cameras are prepared in the rendering server. We assume that $\CamNum$ is a square number because it can efficiently tile the render result into a square texture. A render texture is pre-assigned to each camera as the render target, and $H \times W$ RGB color images, the same size as the input layer of the scoring network, are rendered off-screen in parallel. In the implementation, we set $H = W = 224$.

For each iteration $\Round$, the Python client computes the next exploration space $\Region{\EvalNodeSet}$ and sends the camera parameters, \ie the alignment of center positions and sampled direction vectors ${\DirIdx} (k=1,2,\dots,\DirNum)$, to the rendering server. Based on the camera parameters received, the Unity server moves the cameras according to the camera parameter alignment (\autoref{fig:parallel}, left) and renders $\DirNum$ images in parallel. The contents rendered in $\DirNum$ render textures are then copied and tiled into multiple $H\sqrt{\CamNum} \times W\sqrt{\CamNum}$ render textures (\autoref{fig:parallel}, right). Specifically, the render texture is converted from \texttt{R8G8B8A8\_SRGB} to \texttt{RGB24} using \texttt{AsyncGPUReadback} on Unity and passed to the CPU as a raw byte sequence. This raw byte sequence is sent back to the Python side.

After the Python side converts the received raw byte sequence into a $(\lceil \DirNum / \CamNum \rceil, H\sqrt{\CamNum}, W\sqrt{\CamNum}, 3)$ tensor, this tensor is further transformed into a $(\DirNum, H, W, 3)$ tensor, which is input to the scoring network.

\autoref{fig:parallel} (right) shows the time reduction achieved by using parallel rendering. The results show the average time duration for $\DirNum=1, 12, 36, 72, 144$ over $3\times 10^4$ images and 10 testing scenes, each with 3 trials using different random seeds. Our parallel implementation with $\DirNum=36$ (resp.\,$\DirNum=144$) was 6.0 (resp.\,6.8) times faster than single-threading.

\section{Evaluation}\label{sec:eval}
Since the \productname is divided into the scoring function and the explorer, we evaluate them independently.
All human experiments were ethically reviewed by the Cluster Ethics Review Committee.

\subsection{Comparison of Scoring Network with Humans}
We evaluated whether the scoring network correctly classified human-captured and randomly captured images by comparing them to human classification. In the following, human-captured images will be referred to as positive, and randomly captured images will be referred to as negative.

We asked 75 people to determine whether a given image was taken by a randomly placed camera or taken by a human and processed to remove avatars. The demographics of the respondents were 60 males, 14 females, 1 non-responder, and $32.1 \pm 4.3$ years old. We extracted the images presented to the human from 10 scenes in the test scenes in WI-89, consisting of 100 images in total, with 5 positive and 5 negative images per scene. The same set of images was then classified by the scoring network trained on WI-89.

\begin{table}[t]
    \centering
    \caption{Confusion matrices for the classification accuracy of the humans and the scoring network for the dataset WI-89.}
    \begin{tabular}{cccccc}
        \toprule
        \multirow{2}{*}{Model} & TP & FN & FP & TN & Accuracy \\
        & (\%$\uparrow$) & (\%$\downarrow$)& (\%$\downarrow$)& (\%$\uparrow$)& (\%$\uparrow$) \\
        \midrule
        Humans & 34.6 & 15.3 & 10.7 & 39.3 & 74.0 $\pm$ 21.7        \\
        Ensemble of & \multirow{2}{*}{43.0} & \multirow{2}{*}{7.0} & \multirow{2}{*}{\textbf{5.0}} & \multirow{2}{*}{\textbf{45.0}} & \multirow{2}{*}{\textbf{88.0}}        \\
        Humans & & & & & \\
        ViT-B/16 & \textbf{47.0} & \textbf{3.0} & 12.0 & 38.0 & 85.0 \\
        \bottomrule
    \end{tabular}    
    \label{tab:eval-scoring-net}
\end{table}

\autoref{tab:eval-scoring-net} shows the results. From \autoref{tab:eval-scoring-net}, the accuracy of ViT-B/16 was 85.0\%, 11\% higher than the average human accuracy of 74.0\%. However, there was a large individual variation in human accuracy of $\pm 21.7 \%$. When the majority vote of all participants' choices was used as the classification result for each image, this ensemble result was 88.0\%. Interestingly, the accuracy of this majority decision was comparable to that of ViT. This suggests that ViT may have removed individual bias from human judgments, resulting in an average choice. Furthermore, the presence of large individual differences in classification, even among humans, indicates that the classification of this dataset is a more difficult problem than the classification of general object recognition.

\subsection{Comparison of Explorer with Baselines}
In this section, we evaluate the PanoTree's explorer with a fixed scoring network trained on WI-245.

\subsubsection{Random Explorer}
\begin{figure}[t]
    \centering
    \includegraphics[trim=0 0 0 38.75,clip,width=\linewidth]{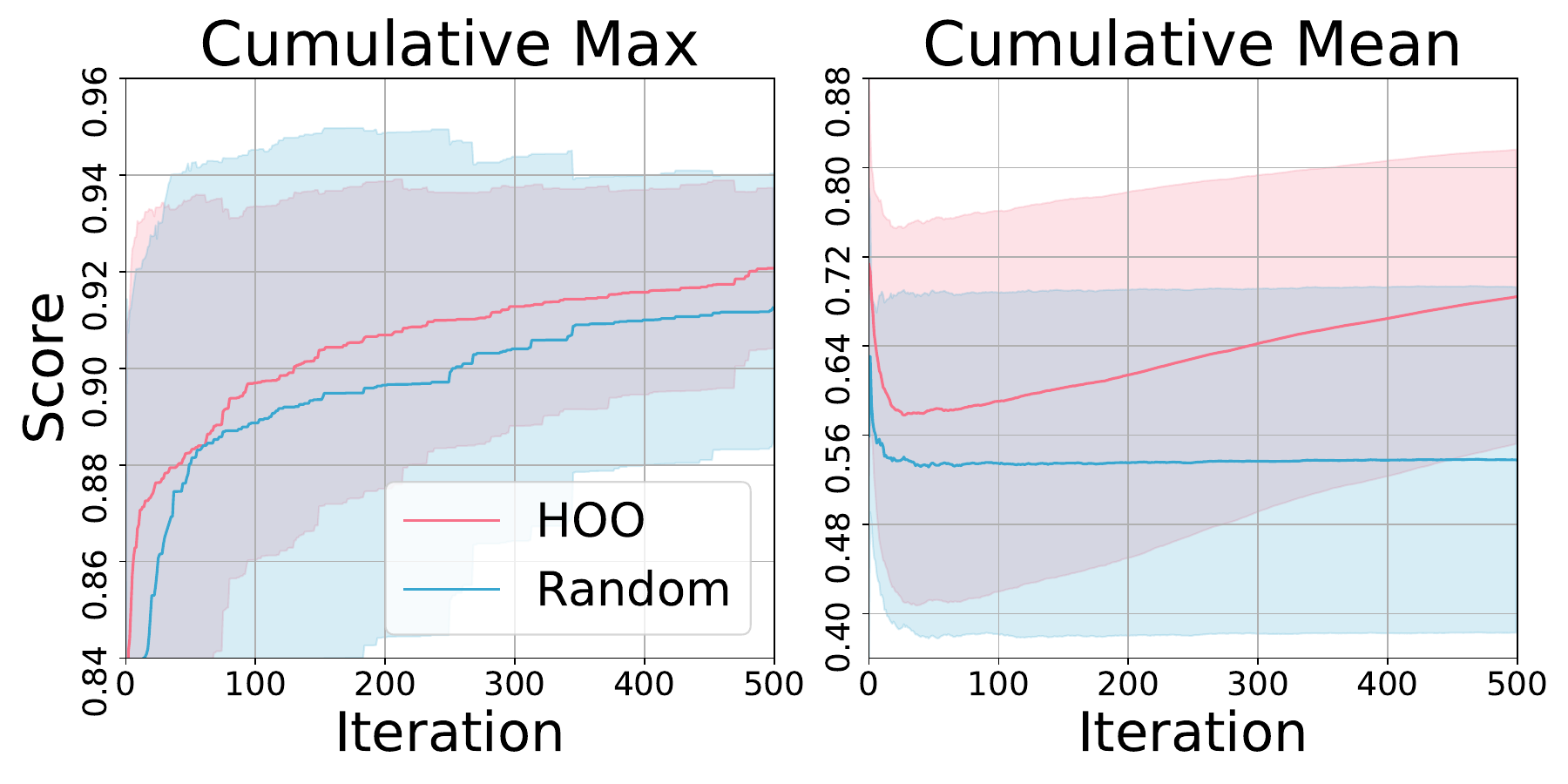}
    \caption{The cumulative maximum score (left) and cumulative average score (right) on WI-245. The results are the average of five trials with different random seeds and ten testing scenes. The filled area indicates the standard deviation of each result. We set $c=0.2, \rho=0.5, v_1=0.5, \DirNum=15$.}
    \label{fig:hoo_vs_random}
\end{figure}

To assess the efficacy of the PanoTree's explorer, we compared it against a naive random search approach. 
The implementation of the random search is as follows: camera coordinates are selected from a three-dimensional uniform distribution within a given bounding box area. Subsequently, the camera's direction vectors are chosen according to \eqref{eq:hoo-direction-sample}, and scores are calculated based on \eqref{eq:reward-by-sampled-dirs}.  For the PanoTree, we set $c=0.2$, $\rho=0.5$, $v_1=0.5$, and for both the random search and the PanoTree, we set $\DirNum=15$.
According to \autoref{tab:eval-exploration}, the PanoTree outperformed the random explorer in both average and maximum scores. Specifically, the PanoTree achieved a maximum score of $0.921 \pm 0.017$ and an average score of $0.684 \pm 0.132$ with both the softmax policy and $D_T=\infty$ included. In contrast, the random method only achieved a maximum score of $0.912 \pm 0.028$ and an average score of $0.538 \pm 0.155$.  \autoref{fig:hoo_vs_random} illustrates that the average score for the random method remained unchanged regardless of the number of iterations. In contrast, the average score for the PanoTree increased with the number of iterations, indicating a continuous improvement over time. These results suggest that the PanoTree prioritizes exploring regions with higher scores more effectively compared to the random method, leading to a more efficient use of the search space.

\begin{table}[t]
    \centering
    \caption{The scores for the exploration algorithms. The results are averaged over five trials with distinct random seeds and ten testing scenes in WI-245.}    
    \begin{tabular}{ccc}
        \toprule
        Explorer & Max Score ($\uparrow$) & Mean Score ($\uparrow$)\\
        \midrule
        Random &  0.912 $\pm$ 0.028 & 0.538 $\pm$ 0.155        \\
        PanoTree  & $\bm{0.921} \pm 0.017$ &  $0.684 \pm 0.132$   \\
        PanoTree (w/o Softmax) &  0.918 $\pm$ 0.021 &  0.655 $\pm$ 0.170 \\
        PanoTree (w/o $D_T=\infty$  ) & $0.911 \pm 0.028$ & $\bm{0.745} \pm 0.097$ \\
        \bottomrule
    \end{tabular}    
    \label{tab:eval-exploration}
\end{table}

\subsubsection{Ablation Study}
\begin{figure}[t]
    \centering
    \includegraphics[trim=0 0 0 38.75,clip,width=\linewidth]{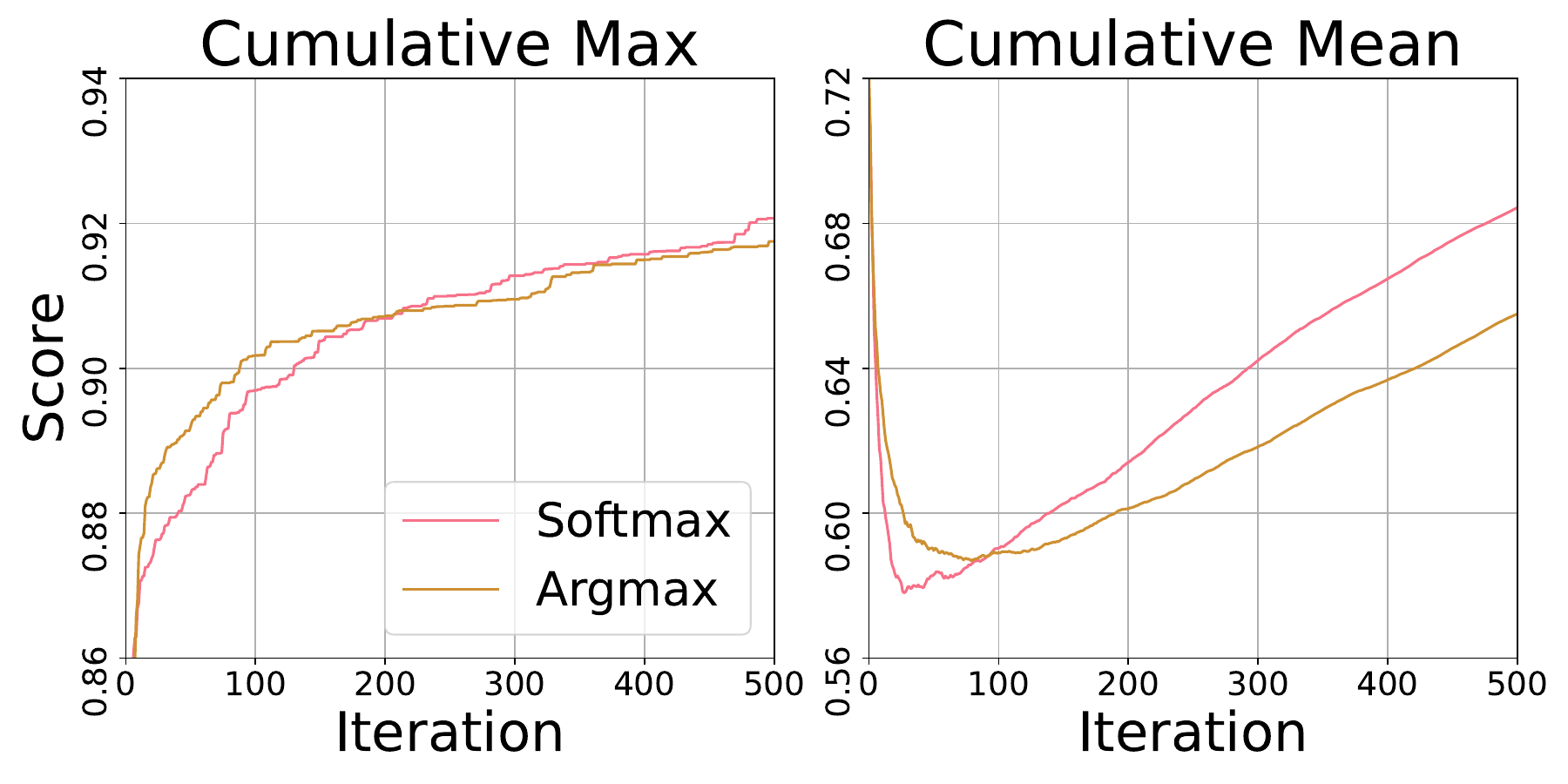}
    \caption{The softmax policy beets the argmax policy. The cumulative maximum score (left) and cumulative average score (right) of the PanoTree with both spatial division policies. The results are the average of five trials with different random seeds and ten testing scenes in WI-245. The standard deviations of softmax cases were less than 0.19, and that of argmax cases were less than 0.22.  We set $c=0.2, \rho=0.5, v_1=0.5, \DirNum=15$.}
    \label{fig:vs_argmax}
\end{figure}

As shown in \autoref{tab:eval-exploration}, using the argmax policy instead of the softmax policy resulted in lower final maximum and average scores compared to the softmax policy. Specifically, the argmax policy achieved a maximum score of $0.918 \pm 0.021$ and an average score of $0.684 \pm 0.132$, which are both lower than the PanoTree scores. According to \autoref{fig:vs_argmax}, the argmax policy performed better in the initial 100 iterations. This is likely because it is beneficial to deterministically divide the space in the early stages of the tree-structured search. However, as iterations progress and the search region narrows, introducing randomness (softmax policy) allows for finer distinctions within regions, enhancing performance. Therefore, combining the argmax policy in the early stages with the softmax policy in the later stages could potentially lead to more efficient exploration.

\autoref{tab:eval-exploration} show that the PanoTree with the depth limit~\eqref{align:depth-limit} achieved the maximum score $0.911 \pm 0.028$ and the mean score $0.745 \pm 0.097$. The mean score was higher than that of the PanoTree with $D_T=\infty$, but the maximum score was lower or comparable with that of the random explorer. Therefore, we do not use the depth limit in subsequent experiments.

\subsection{Scaling Up Experiments}
In this section, we evaluate the extent to which performance improvements can be expected with an increase in the number of tree updates (number of exploration nodes), the number of direction vectors, and the number of training scenes.

\subsubsection{Enlarging Dataset}
This study's primary focus is exploring unknown scenes, emphasizing the importance of the scoring network's generalization performance across scenes. Therefore, this section investigates how increasing the number of training scenes affects the scoring network's performance.
The dataset WI-89 was limited to 89 scenes, with 79 used for training. In contrast, the dataset WI-245 is approximately three times larger, encompassing 245 scenes, with 235 scenes dedicated to training (\autoref{tab:dataset}).
A pre-trained ViT-B/16 with the same parameters was employed to compare the two datasets, using the same hyperparameters for model training except for the learning rate. Experiments were conducted with base learning rates of $1\times10^{-5},2\times 10^{-5},4\times 10^{-5},8 \times 10^{-5}$ and the results for the learning rate yielding the highest average test accuracy, $4 \times 10^{-5}$, are presented in \autoref{tab:increasing_datasets}. 
Our study's key finding, as indicated in \autoref{tab:increasing_datasets}, is that increasing the number of training scenes and training images by approximately four times led to a significant 13.29\% improvement in test accuracy. This finding underscores the direct and positive impact of increasing the number of training scenes on scoring accuracy.

\begin{table}[t]
    \centering
    \caption{Test accuracy of the scoring network ViT-B/16 trained on two types of datasets, WI-89 and WI-245, along with the number of scenes and images used for training the network in each dataset. The results represent the average of eight trials, each with different random seeds.}
    \begin{tabular}{cccc}
    \toprule
    Dataset & Train Scenes & Train Images & Test Accuracy \\
    \hline
    WI-89 & 79  & $0.4 \times 10^6$ & $75.99 \pm 0.52$ \\
    WI-245 & 235 & $1.6 \times 10^6$ & $\mathbf{88.28} \pm 0.27$\\
    \hline
    \end{tabular}
    \label{tab:increasing_datasets}
\end{table}

\subsubsection{Increasing the Time Horizon $\EpochNum$}

\begin{figure}[t]
    \centering
    \includegraphics[trim=0 0 0 38.75, clip, width=\linewidth]{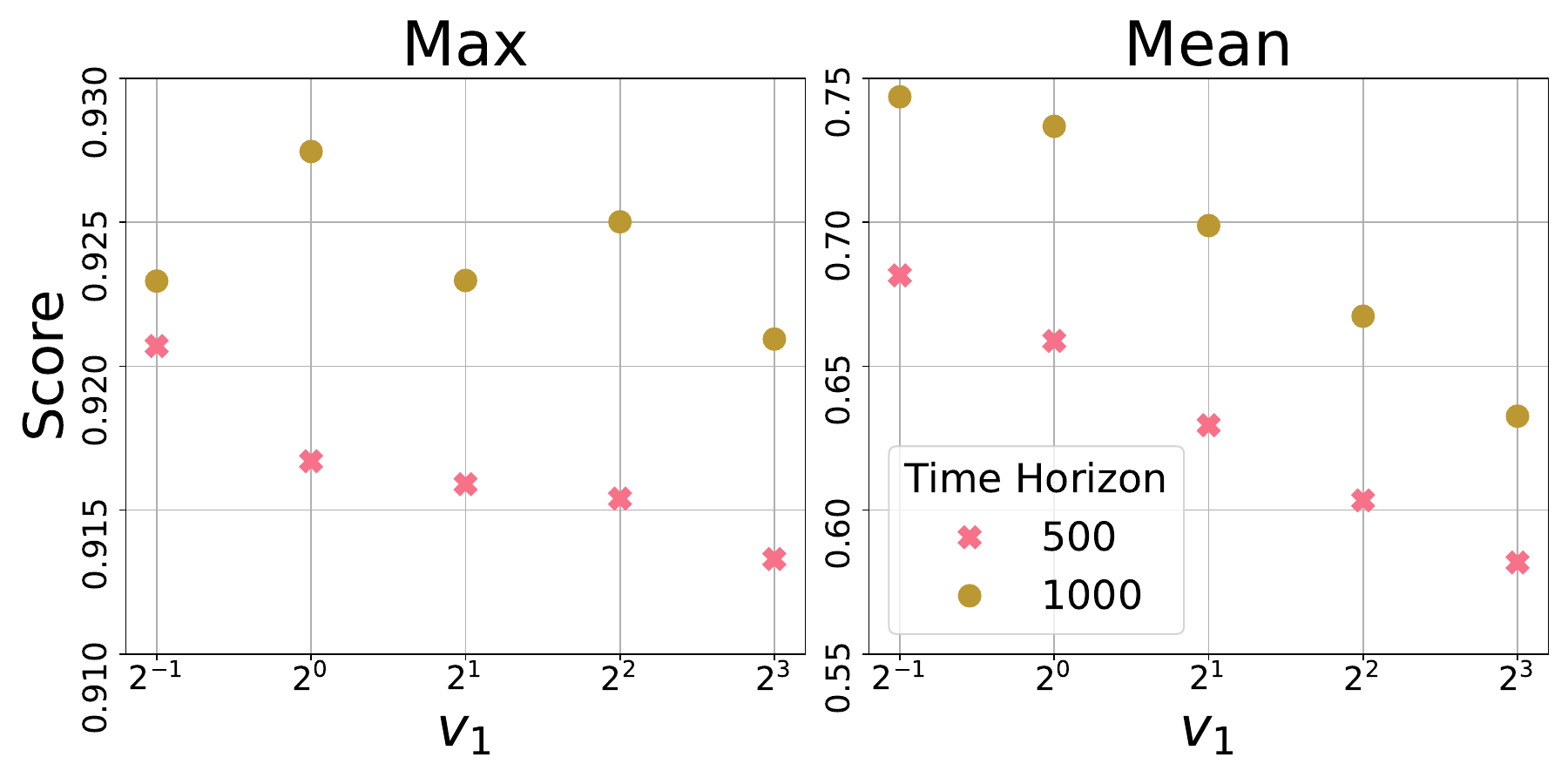}
    \caption{ As the time horizon increased from 500 to 1000, both the maximum score (left) and the mean score (right)  increased. The results hold for all values of $v_1$ in $2^{n} (n=-1, 0, \dots, 3)$.
    The results are averaged over five trials with different random seeds and ten testing scenes in WI-245. The standard deviation for maximum (resp.\,mean) was less than 0.026 (resp.\,0.16).
    }
    \label{fig:increasing_updates_vs_v1}
\end{figure}

\autoref{fig:increasing_updates_vs_v1} presents the improvements in scores in response to increasing the number of iterations $\EpochNum$ from 500 to 1000. Here, $\rho=0.5$ and $\DirNum=15$ are fixed, and experiments were conducted for $v_1=2^{n} (n=-1, 0, \dots, 3)$.
Specifically, for $\EpochNum=500$, $v_1=2^{-1}$ was optimal for the maximum score, while for $\EpochNum=1000$, $v_1=1$ was optimal. Here, a larger $v_1$ means exploring a wider area in the early stages of the search. Therefore, the increase in $\EpochNum$ allowed for more extensive exploration, which aligned well with a larger $v_1$, resulting in higher performance.

\subsubsection{Increasing Directions  $\DirNum$ }

\begin{figure}[t]
    \centering
    \includegraphics[trim=0 0 0 38.75, clip,width=\columnwidth]{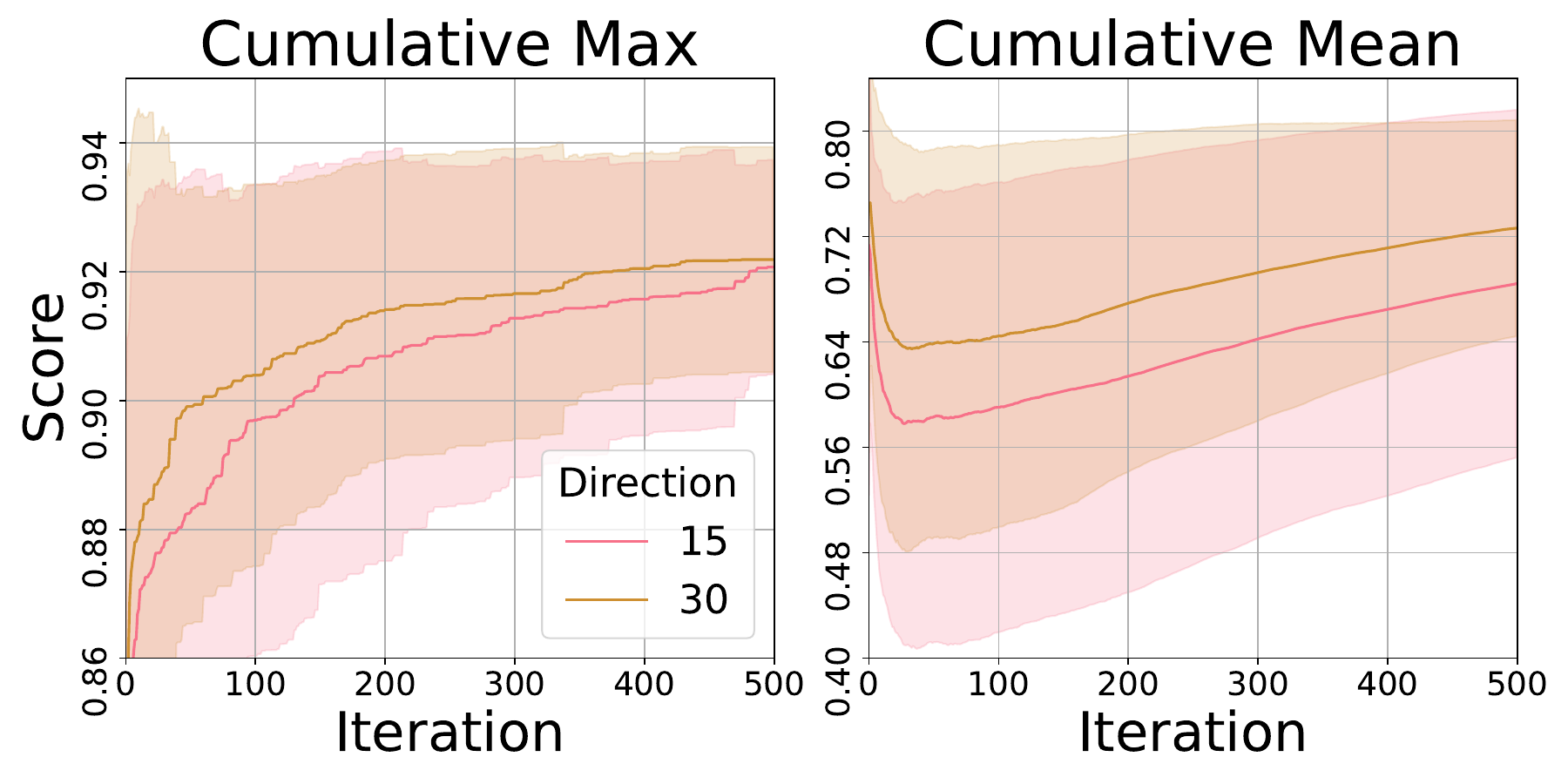}
    \caption{Cumulative maximum scores (left) and cumulative mean scores (right) for $\DirNum=15,30$.  The results are averaged over five trials with different random seeds and ten testing scenes in WI-245. 
    }
    \label{fig:incresing_dir_vs30}
\end{figure}

\autoref{fig:incresing_dir_vs30} illustrates the changes in the maximum score when the number of direction vectors is increased from 15 to 30. We set \(\EpochNum=500\), \(c=0.2\), and \(\rho=v_1=0.5\).
At the last iteration, 
the max score  $0.922 \pm 0.017$ with the mean score  $0.726 \pm 0.082$  when $\DirNum=30$. 
However, when \(\DirNum\) increased from 15 to 30, the increment was smaller than 0.042, which is less than the growth for a comparable increase in \(\EpochNum\) from 500 to 1000 in \autoref{fig:increasing_updates_vs_v1}. 
These results suggest that depth-first search is more significant in our scene exploration.  

\section{Application}\label{sec:appliaction}
The \productname~has the potential to be applied to a wide range of applications, including the design of VR spaces and the enhancement of the user experience within these spaces. This section introduces potential applications of \productname.
Please refer to the supplementary video for examples of these applications in operation.

\subsection{Photospot Visualization of VR worlds in Production}
\autoref{fig:teaser} (c) shows the visualization of photo spots in a  VR world. By visualizing the photo spots, users can easily identify the optimal position and direction for taking a "good" photo, enhancing the user experience as in real-world tourism. 
Additionally, for VR world designers, the visualization provides valuable feedback on the quality of their creations. By repeatedly creating VR scenes and evaluating them with \productname, the quality of the VR world can be iteratively improved. 

In the future, if the properties of all objects in VR scenes can be defined in a differentiable state, such as their position, angle, and appearance, it would be possible to optimize these properties so that the scene's overall score is high.

\subsection{Automatic Thumbnail Generation}
As shown in \autoref{fig:teaser} (d), the high-scoring photos from \productname~can be used for the automated generation of thumbnails for VR worlds. On social VR platforms, VR world designers can upload images that best represent their VR worlds as thumbnails. The creation of more attractive thumbnails is expected to result in increased user engagement with the VR world. While online video-sharing services offer automatic thumbnail generation, the majority of social VR platforms lack this functionality, making it a time-consuming task for each VR designer.

\productname~presents the designer with a selection of high-scoring images from the explored photo spots in the VR scene. The VR designer can then select one of these images and register it as a thumbnail. In the future, the scoring network will be improved by human-in-the-loop training, whereby the designer's selection will be recorded, and the scoring network will be updated accordingly.

\subsection{Visitor Circulation Planning}
By following the multiple photo spots found in the \productname, a visitor circulation plan within the VR world can be proposed (see supplementary video).
Since visitors can enter and exit the VR world at any time, they tend to leave the world immediately, lacking guidance on what to do when they enter the VR world. This application can be used to give visitors the task of visiting photo spots as if they were on a real-life tour, which is expected to increase the time visitors spend in the space. Furthermore, if the VR world creator can establish a flow line that allows visitors to tour the entire VR space, they will be able to experience the experience design intended by the creator and more effectively convey the appeal of the production.

\section{Limitation and Future Work}

\subsection{Differentiable Rendering}\label{sec:discuss:nerf}
When exploring the best camera direction for each space, the current method samples multiple camera directions and selects the direction with the highest reward. This is because rendering by game engines is non-differentiable and cannot calculate the change in reward when the camera direction is slightly changed.
Recently, methods such as NeRF~\cite{mildenhall2020nerf} and Gaussian splatting~\cite{kerbl2023gaussian} have been proposed to represent entire 3D scenes in a differentiable form, such as neural networks and Gaussian distributions. If the entire VR scene can be reconstructed with such differentiable representations, the derivative of the reward can be back-propagated to the camera position and direction.
This would allow for more precise optimization of the direction vectors to obtain good images and further speed up the search process. 
Differentiable optimization that includes the direction and position of the camera has already been proposed in neural rendering~\cite{lin2021barf, park2023camp} and is expected to be applied in this study.

\subsection{Deep Policy Network}
In this study, deep learning is only used for the scoring function, and the region division policy is implemented using a softmax function based on the dimensions of the bounding box. Replacing this policy function with a neural network could be valuable in reflecting peripheral visual information and enabling smarter division. 
Deep policy networks appear in \cite{silver2017mastering} for the Monte Carlo tree search (MCTS). 
By replacing MCTS with HOO, deep policy networks can be applied to the PanoTree.
Our original softmax policy has the advantage of being computationally fast and not requiring policy training compared to deep methods. Investigating the trade-off between speed and performance of these methods would be an interesting direction for future work.

\section{Conclusion}
This paper introduces the PanoTree, an autonomous photo-spot explorer for 3D scenes of VR worlds. To realize our photo-spot exploration, we propose a framework that combines a deep photo-scoring network for determining the likelihood of images being taken by humans and a spatial exploration algorithm based on HOO. Our scoring network, trained on millions of human-captured screenshots within VR platforms, achieved a level of accuracy comparable to human evaluation in determining whether an input image was taken by a human.

Moreover, our exploration algorithm achieved a 6.0 times speedup through parallel exploration and outperformed the random explorer by 33\% in mean scores by introducing a probabilistic exploration strategy to the traditional HOO. Further experiments with the explorer revealed the importance of scaling up iterations, directions, and image datasets. Finally, we proposed applications leveraging PanoTree, including automatic thumbnail suggestions for VR worlds, visualization of photo spots, and tour route planning for VR world scenes.

Our approach demonstrates that the vast amount of user activity data accumulated on VR platforms is generating new datasets and insights. As human life gradually shifts towards virtual spaces where all actions can be recorded, this indicates the possibility of gaining a new understanding of humans and the spaces we inhabit, transcending the boundaries between virtual and real. We believe that the large-scale data-driven understanding of humans and spaces in VR platforms proposed by this research will inspire future researchers.

%% file: appendix.tex
\renewcommand{\thesection}{\Alph{section}}
\renewcommand{\theequation}{S.\arabic{equation}}
\renewcommand{\thefigure}{S.\arabic{figure} }
\renewcommand{\thetable}{S.\arabic{table} }

\setcounter{equation}{0}
\setcounter{figure}{0}
\setcounter{table}{0}

\begin{center}
{\sffamily\ifvgtcjournal\huge\else\LARGE\bfseries\fi%
   Supplementary Material \\
   {\LARGE
   for \productname
   }
}
\end{center}

\section{Summary of Notations}
\autoref{tab:notaions} summarize the notations used in the paper. In this paper, we frequently use the notation \emph{region}. Here, a region means a cuboid with or without faces. The presence or absence of faces in the region does not affect the PanoTree since only the region's center position and size are needed for the algorithm. Furthermore, our definition of the region is consistent with the requirement in the HOO \cite{bubeck2011xarmed} to be a measurable subset of $\RegionAll$.

\begin{table}[ht]
    \centering
    \caption{Notation Description}
    \begin{tabular}{c|l}
    \toprule
    Notation &  Description\\
    \hline
       $\R$ & The set of real numbers.\\

       $H, W, C$ & The height, the width, and the number of channels of input images for scoring functions. \\
       
       $\RegionAll$ & The entire region of exploration determined by collisions. We have $\RegionAll \subset \R^3$ except for the 1D-example.\\

        $x,y,z$ & The indexes of axes in the 3D space. The $y$ represents the height axis. \\
 
        $L_x, L_y, L_z$  & The length of a region's edge corresponding axis.\\

        $f_\mathrm{R}$ & The rendering function.\\

        $f_\mathrm{S}$ & The scoring network.\\

        $\CamNum$ & The number of cameras used in the tile rendering.\\

        $\DirNum$ & The number of camera directions per position.\\

        $\DirIdx$  & The camera direction for the index $k$. ($k=0, 1, \dots, \DirNum-1$.)\\

        $\EpochNum$ & The number of exploration iterations.\\

       $\Round$  &  The index of the exploration iteration. ($\Round=0,1, \dots, \EpochNum-1.$)\\

        $c$ & A hyper-parameter of the  HOO, representing the power of exploration regularization. ($c>0$)\\

        $v_1$ & A hyper-parameter of the HOO, representing the power of the spatial regularization. ($v_1 > 0$.)\\

        $\rho$ & A hyper-parameter of the HOO, representing the reduction of the spatial regularization according to the depth $\Depth$. ($0<\rho <1 $.) \\

        $\SubtreeIdx{n}$ & The constructed tree at the iteration $n$.\\

        $D_{\EpochNum}$ & The upper bound of the tree.\\

       $(\NodeSet)$  &  A node in the tree, where $\Depth$ is the depth of the node and  $\NodeIdx=0, 1, \dots, 2^{d}-1$  is the index of the node in depth $\Depth$.\\

       $\Region{\Depth,\NodeIdx}$ & The region contained in $\RegionAll$ corresponding to the node $(\NodeSet)$.\\ 

        $\Vector{c}_{\NodeSet}$ & The center of $\Region{\Depth,\NodeIdx}$.\\

       $(\EvalNodeSet)$  & The selected node in the iteration $n$ to be evaluated.\\

      $\ScoreN$ & The reward in the $\Round$-th iteration.\\

       $T_{d,i}(n)$ &  The number of searches of        $(\NodeSet)$ up to and including iteration $\Round$. ($\forall (\NodeSet), \Visit{\NodeSet}(0)=0$.)\\

       $\AvgScoreDef$  & The average reward of the node $(\NodeSet)$ in  $\Round$-th iteration. ($\forall (\NodeSet), \AvgScore{\NodeSet}(0)=0$.)  \\
       
       $U_{d,i}(n)$ & The upper bound estimation of the node $(d,i)$ in $n$-th iteration.  ($\forall (\NodeSet), \UVal{\NodeSet}(0)=+\infty$.)\\
       
       $B_{d,i}(n)$  & The refined upper bound estimation  of the node $(d,i)$ in $n$-th iteration. ($\forall (\NodeSet), \BVal{\NodeSet}(0)=+\infty$.) \\

    \end{tabular}
    \label{tab:notaions}
\end{table}

\section{Detail of Datasets}
\autoref{tab:dataset_detail} shows the detailed statistics of scene and image datasets of WI-89 and WI-245.  The test scenes of both WI-89 and WI-245 are mutually distinct and separated from all train scenes of both datasets.

\begin{table}[ht]
    \centering
    \caption{Detailed Statistics of  WI-89 (left) and  WI-245 (right). }
    \begin{tabular}{c|ccccc}
    \toprule
        Split &  Scenes &  Positive Imgs. & Negative Imgs. & Total Imgs.\\
       \midrule
        Train & 79 & $1.86 \times 10^5$  &  $1.72 \times 10^5$  &  $3.59 \times 10^5$ \\
        Test  & 10 &  $2.03 \times 10^4$  &   $2.16 \times 10^4$ &   $4.19 \times 10^4$ \\     
        Total & 89 &  $2.07 \times 10^5$ &  $1.94 \times 10^5$ &  $4.01 \times 10^5$
    \end{tabular}
        \begin{tabular}{c|ccccc}
        \toprule
        Split &  Scenes &  Positive Imgs. & Negative Imgs. & Total Imgs.\\
       \midrule
        Train & 235 & $7.95 \times 10^5$  &  $7.92 \times 10^5$  &  $1.59 \times 10^6$ \\
        Test  & 10 &  $3.14 \times 10^4$  &   $3.20 \times 10^4$ &  $6.34 \times 10^4$ \\     
        Total & 245 & $8.27 \times 10^5$ &   $8.24 \times 10^5$ &  $1.65 \times 10^6$
    \end{tabular}

    \label{tab:dataset_detail}
\end{table}

\section{Experimental Settings of Finetuning}
To improve the scoring network's generalization performance, we applied data augmentation techniques to our dataset. Since our dataset consists of CG data, it is unclear whether existing data augmentation methods used for ImageNet-1k, which consists of real photos, can be effectively applied to our dataset. Therefore, as a preliminary experiment, we investigated effective data augmentation methods for our training.

\begin{figure}[th]
    \centering
    \includegraphics[width=0.33\columnwidth]{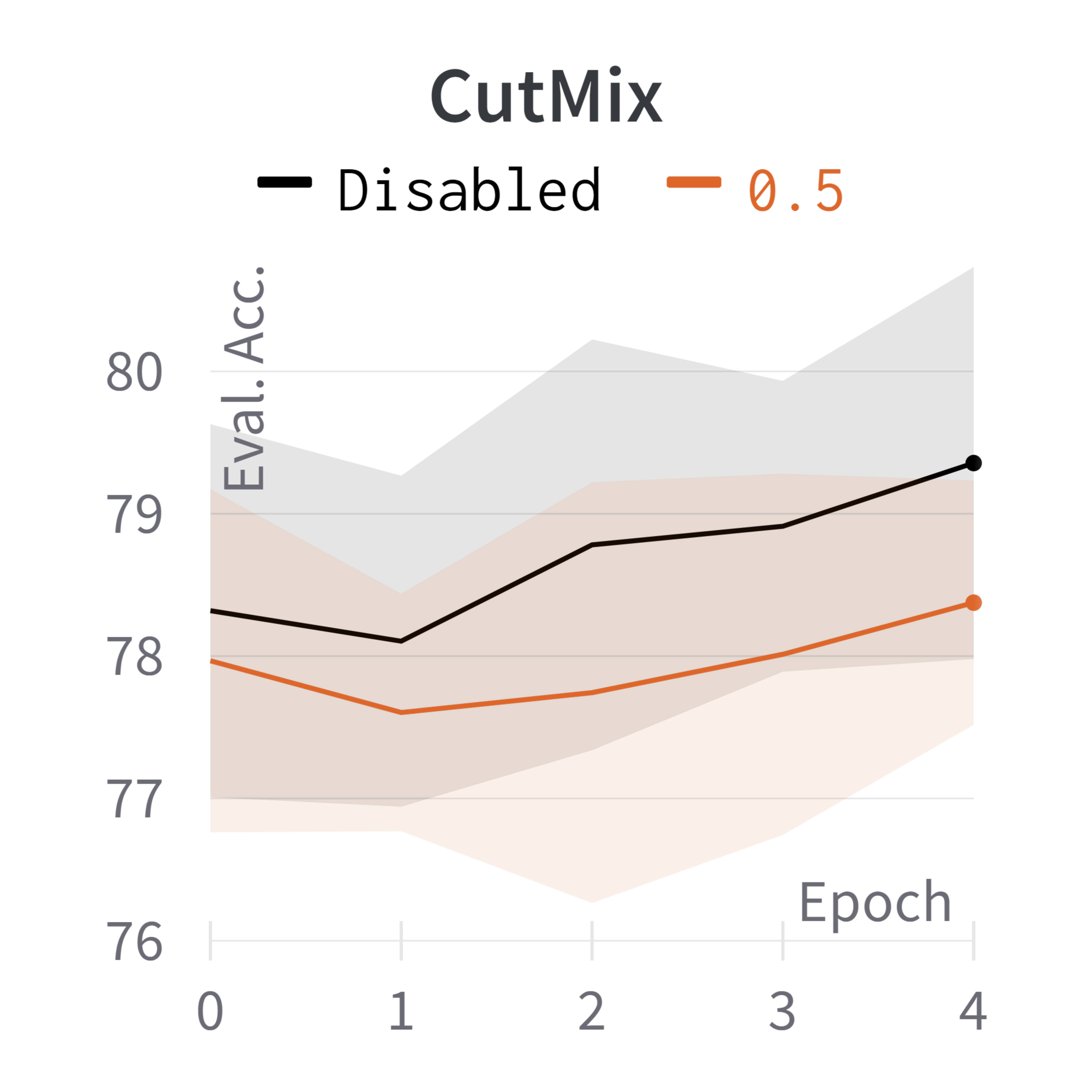}        \includegraphics[width=0.33\columnwidth]{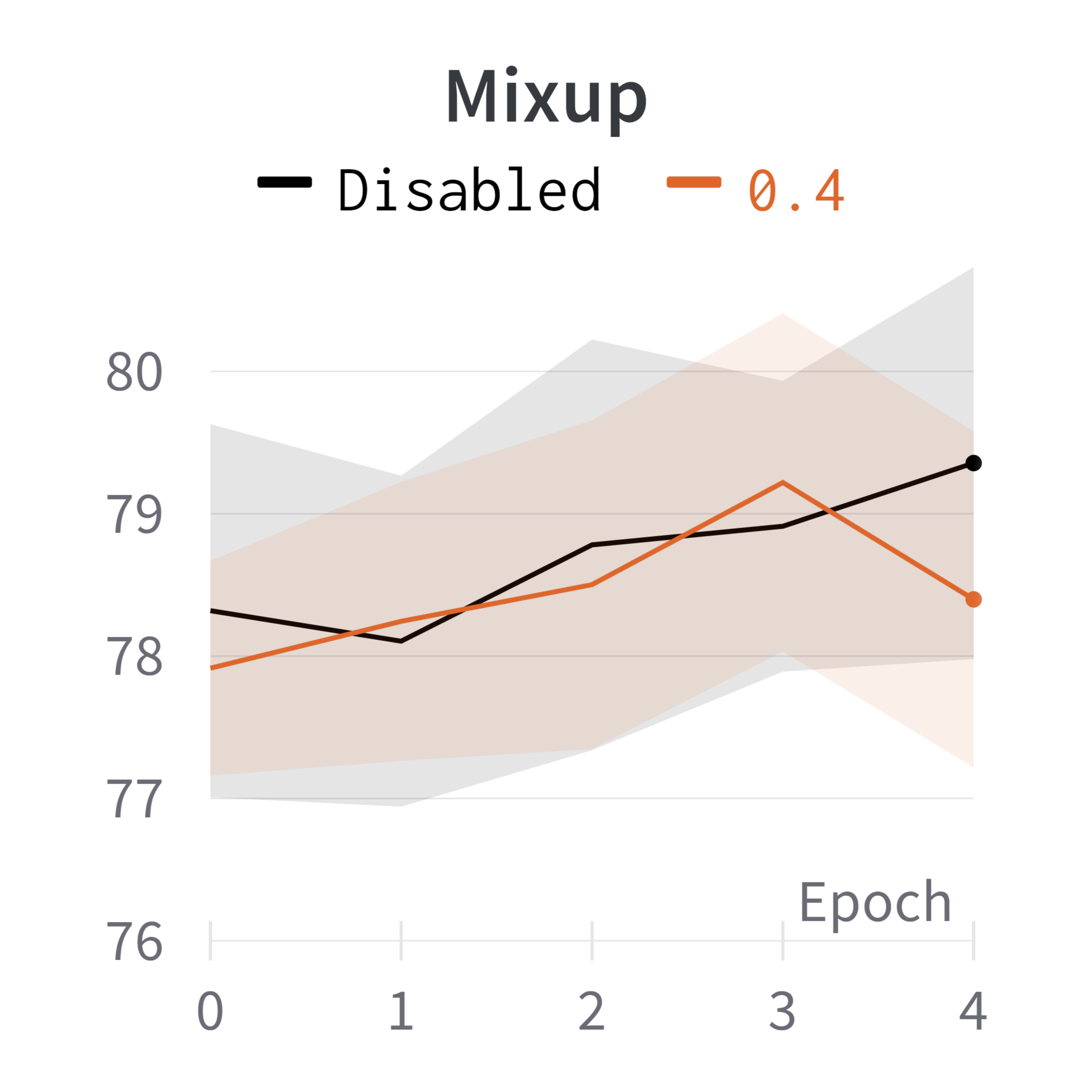}
    \caption{Epochs and accuracy when training with (left) CutMix~=~0.5 and (right) Mixup~=~0.4 enabled (orange line) and disabled (black line) on ViT-B/16. 
    }
    \label{fig:dataaug-comparison}
\end{figure}

\autoref{tab:aug_hps} shows the hyperparameters for the training. 
As shown in \autoref{fig:dataaug-comparison}, the classification performance of the scoring network decreased when CutMix~\cite{yun2019cutmix} and Mixup~\cite{zhang2017mixup} were enabled, which are effective for ImageNet.
Since these methods paste or blend two images, we confirmed that these methods are not effective for our dataset, as they significantly collapse the image features.
Therefore, we do not use the CutMix and the Mixup.
From the above, we used only Random Erasing~\cite{zhong2020random} set to 0.1, and we used RandAugment~\cite{cubuk2020randaugment} $(m=9, n=5)$ in which the option of increasing the severity of the augmentations with magnitude (inc) was set to True.

\begin{table}[th]
    \centering
    \caption{Summary of hyperparameters for fine-tuning scoring networks.}
    \begin{tabular}{c|c}
    \toprule
    Params & Values\\
    \midrule
    Epochs & 5\\
    Mini-batch size     & 256  \\
    Base learning rate & $10^{-5}$\\
    Schedule & Cosine annealing \\
    Warmup epochs & 1 \\
    Warmup learning rate &  $10^{-6}$\\
    Dropout     &  0.5\\
    RandAugument \cite{cubuk2020randaugment} & $m=9, n=5, \text{inc}=\text{True}$\\
    Random Erasing \cite{zhong2020random} & 0.1\\
    Label Smoothing \cite{muller2019does} & 0.1\\
    CutMix~\cite{yun2019cutmix}   & Disabled\\
    Mixupp~\cite{zhang2017mixup} & Disabled \\
    Random seeds & Eight different seeds\\
    \hline
    \end{tabular}
    \label{tab:aug_hps}
\end{table}

%% file: main.bbl
\begin{thebibliography}{10}

\bibitem{Peter2003ucb}
P.~Auer.
\newblock Using confidence bounds for exploitation-exploration trade-offs.
\newblock {\em The Journal of Machine Learning Research}, 3(3):397^^e2^^80^^93422, Mar 2003.

\bibitem{angeles2022social}
M.~Barreda-^^c3^^81ngeles and T.~Hartmann.
\newblock Psychological benefits of using social virtual reality platforms during the covid-19 pandemic: The role of social and spatial presence.
\newblock {\em Computers in Human Behavior}, 127:107047, 2022.

\bibitem{brochu2010tutorial}
E.~Brochu, V.~M. Cora, and N.~De~Freitas.
\newblock A tutorial on bayesian optimization of expensive cost functions, with application to active user modeling and hierarchical reinforcement learning.
\newblock {\em arXiv preprint arXiv:1012.2599}, 2010.

\bibitem{bubeck2011xarmed}
S.~Bubeck, R.~Munos, G.~Stoltz, and C.~Szepesvari.
\newblock X-armed bandits.
\newblock {\em arXiv preprint arXiv:1012.2599}, 2011.

\bibitem{sliwecki2021virtual}
B.~^^c5^^9aliwecki.
\newblock Virtual reality architectural spaces and the shift of populace in online social vr platforms in 2020.
\newblock {\em Architecturae et Artibus}, 13(4):1--12, 2021.

\bibitem{cluster2023}
{C}luster {I}nc.
\newblock "{M}etaverse {P}latform - {C}luster," https://cluster.mu/en (accessed {J}an. 29, 2024).

\bibitem{cubuk2020randaugment}
E.~D. Cubuk, B.~Zoph, J.~Shlens, and Q.~V. Le.
\newblock Rand{A}ugment: Practical automated data augmentation with a reduced search space.
\newblock In {\em Proceedings of the IEEE/CVF conference on computer vision and pattern recognition workshops (CVPR)}, pages 702--703, 2020.

\bibitem{deng2017image}
Y.~Deng, C.~C. Loy, and X.~Tang.
\newblock Image aesthetic assessment: An experimental survey.
\newblock {\em IEEE Signal Processing Magazine}, 34(4):80--106, 2017.

\bibitem{dosovitskiy2021an}
A.~Dosovitskiy, L.~Beyer, A.~Kolesnikov, D.~Weissenborn, X.~Zhai, T.~Unterthiner, M.~Dehghani, M.~Minderer, G.~Heigold, S.~Gelly, J.~Uszkoreit, and N.~Houlsby.
\newblock An image is worth 16x16 words: Transformers for image recognition at scale.
\newblock In {\em International Conference on Learning Representations}, 2021.

\bibitem{eiben2015introduction}
A.~E. Eiben and J.~E. Smith.
\newblock {\em Introduction to evolutionary computing}.
\newblock Springer, 2015.

\bibitem{Hu2019graphbased}
C.~Y. Fei~Hu, Zhenlong~Li and Y.~Jiang.
\newblock A graph-based approach to detecting tourist movement patterns using social media data.
\newblock {\em Cartography and Geographic Information Science}, 46(4):368--382, 2019.

\bibitem{garcia2018survey}
A.~Garcia-Garcia, S.~Orts-Escolano, S.~Oprea, V.~Villena-Martinez, P.~Martinez-Gonzalez, and J.~Garcia-Rodriguez.
\newblock A survey on deep learning techniques for image and video semantic segmentation.
\newblock {\em Applied Soft Computing}, 70:41--65, 2018.

\bibitem{giunchi2024dreamcode}
D.~Giunchi, N.~Numan, E.~Gatti, and A.~Steed.
\newblock {D}ream{C}ode{VR}: Towards democratizing behavior design in virtual reality with speech-driven programming.
\newblock In {\em Proceedings of IEEE International Conference of Virtual Reality 2024}, pages 579--589, 2024.

\bibitem{gosal2019social}
A.~S. Gosal, I.~R. Geijzendorffer, T.~V^^c3^^a1clav^^c3^^adk, B.~Poulin, and G.~Ziv.
\newblock Using social media, machine learning and natural language processing to map multiple recreational beneficiaries.
\newblock {\em Ecosystem Services}, 38:100958, 2019.

\bibitem{Hai2018-jl}
W.~Hai, N.~Jain, A.~Wydra, N.~M. Thalmann, and D.~Thalmann.
\newblock Increasing the feeling of social presence by incorporating realistic interactions in multi-party {VR}.
\newblock In {\em Proceedings of the 31st International Conference on Computer Animation and Social Agents}, pages 7--10, May 2018.

\bibitem{hosu2019effective}
V.~Hosu, B.~Goldlucke, and D.~Saupe.
\newblock Effective aesthetics prediction with multi-level spatially pooled features.
\newblock In {\em proceedings of the IEEE/CVF conference on computer vision and pattern recognition (CVPR)}, pages 9375--9383, 2019.

\bibitem{Hu2015extracting}
Y.~Hu, S.~Gao, K.~Janowicz, B.~Yu, W.~Li, and S.~Prasad.
\newblock Extracting and understanding urban areas of interest using geotagged photos.
\newblock {\em Computers, Environment and Urban Systems}, 54:240--254, 2015.

\bibitem{huang2023global}
H.~Huang, J.~Qiu, and K.~Riedl.
\newblock On the global convergence of particle swarm optimization methods.
\newblock {\em Applied Mathematics \& Optimization}, 88(2):30, 2023.

\bibitem{Kaimal2020-je}
G.~Kaimal, K.~Carroll-Haskins, M.~Berberian, A.~Dougherty, N.~Carlton, and A.~Ramakrishnan.
\newblock Virtual reality in art therapy: A pilot qualitative study of the novel medium and implications for practice.
\newblock {\em Art Therapy}, 37(1):16--24, Jan. 2020.

\bibitem{kerbl2023gaussian}
B.~Kerbl, G.~Kopanas, T.~Leimk{\"u}hler, and G.~Drettakis.
\newblock 3d gaussian splatting for real-time radiance field rendering.
\newblock {\em ACM Transactions on Graphics}, 42(4), July 2023.

\bibitem{li2018bigdata}
J.~Li, L.~Xu, L.~Tang, S.~Wang, and L.~Li.
\newblock Big data in tourism research: A literature review.
\newblock {\em Tourism Management}, 68:301--323, 2018.

\bibitem{lin2023magic3d}
C.-H. Lin, J.~Gao, L.~Tang, T.~Takikawa, X.~Zeng, X.~Huang, K.~Kreis, S.~Fidler, M.-Y. Liu, and T.-Y. Lin.
\newblock Magic3d: High-resolution text-to-3d content creation.
\newblock In {\em IEEE Conference on Computer Vision and Pattern Recognition ({CVPR})}, 2023.

\bibitem{lin2021barf}
C.-H. Lin, W.-C. Ma, A.~Torralba, and S.~Lucey.
\newblock {BARF}: Bundle-adjusting neural radiance fields.
\newblock In {\em IEEE International Conference on Computer Vision ({ICCV})}, 2021.

\bibitem{ma2018language}
R.~Ma, A.~G. Patil, M.~Fisher, M.~Li, S.~Pirk, B.-S. Hua, S.-K. Yeung, X.~Tong, L.~Guibas, and H.~Zhang.
\newblock Language-driven synthesis of 3d scenes from scene databases.
\newblock In {\em ACM SIGGRAPH Asia 2018 Technical Papers}, page 212, 2018.

\bibitem{Makransky2018-nb}
G.~Makransky and L.~Lilleholt.
\newblock A structural equation modeling investigation of the emotional value of immersive virtual reality in education.
\newblock {\em Springer Educational Technology Research and Development}, 66(5):1141--1164, Oct. 2018.

\bibitem{Joshua2019shaping}
J.~McVeigh-Schultz, A.~Kolesnichenko, and K.~Isbister.
\newblock Shaping pro-social interaction in {VR}: An emerging design framework.
\newblock In {\em Proceedings of the 2019 CHI Conference on Human Factors in Computing Systems (CHI '19)}, page 1^^e2^^80^^9312, New York, NY, USA, 2019.

\bibitem{Merrell2011furniture}
P.~Merrell, E.~Schkufza, Z.~Li, M.~Agrawala, and V.~Koltun.
\newblock Interactive furniture layout using interior design guidelines.
\newblock In {\em ACM SIGGRAPH 2011 Papers}, SIGGRAPH '11, New York, NY, USA, 2011. Association for Computing Machinery.

\bibitem{mildenhall2020nerf}
B.~Mildenhall, P.~P. Srinivasan, M.~Tancik, J.~T. Barron, R.~Ramamoorthi, and R.~Ng.
\newblock Ne{RF}: Representing scenes as neural radiance fields for view synthesis.
\newblock In {\em Proceedings of European Conference on Computer Vision}, 2020.

\bibitem{minaee2021image}
S.~Minaee, Y.~Boykov, F.~Porikli, A.~Plaza, N.~Kehtarnavaz, and D.~Terzopoulos.
\newblock Image segmentation using deep learning: A survey.
\newblock {\em IEEE Transactions on Pattern Analysis and Machine Intelligence}, 44(7):3523--3542, 2021.

\bibitem{muller2019does}
R.~M{\"u}ller, S.~Kornblith, and G.~E. Hinton.
\newblock When does label smoothing help?
\newblock {\em Advances in Neural Information Processing systems}, 32, 2019.

\bibitem{nelder1965simplex}
J.~A. Nelder and R.~Mead.
\newblock A simplex method for function minimization.
\newblock {\em The Computer Journal}, 7(4):308--313, 1965.

\bibitem{park2023camp}
K.~Park, P.~Henzler, B.~Mildenhall, J.~T. Barron, and R.~Martin-Brualla.
\newblock Camp: Camera preconditioning for neural radiance fields.
\newblock {\em ACM Transactions on Graphics}, 2023.

\bibitem{poole2023dreamfusion}
B.~Poole, A.~Jain, J.~T. Barron, and B.~Mildenhall.
\newblock Dreamfusion: Text-to-3d using 2d diffusion.
\newblock In {\em the Eleventh International Conference on Learning Representations (ICLR)}, 2023.

\bibitem{powell1964efficient}
M.~J. Powell.
\newblock An efficient method for finding the minimum of a function of several variables without calculating derivatives.
\newblock {\em The Computer Journal}, 7(2):155--162, 1964.

\bibitem{Oliveira2023Virtual}
T.~a. Ribeiro~de Oliveira, B.~Biancardi~Rodrigues, M.~Moura~da Silva, R.~Antonio N.~Spinass\'{e}, G.~Giesen~Ludke, M.~Ruy Soares~Gaudio, G.~Iglesias Rocha~Gomes, L.~Guio~Cotini, D.~da~Silva~Vargens, M.~Queiroz~Schimidt, R.~Varej\~{a}o Andre\~{a}o, and M.~Mestria.
\newblock Virtual reality solutions employing artificial intelligence methods: A systematic literature review.
\newblock {\em ACM Computing Survey}, 55(10), feb 2023.

\bibitem{roberts2022steps}
J.~Roberts, A.~Banburski-Fahey, and J.~Lanier.
\newblock Steps towards prompt-based creation of virtual worlds.
\newblock {\em arXiv preprint arXiv:2211.05875}, 2022.

\bibitem{roberts1963machine}
L.~G. Roberts.
\newblock {\em Machine perception of three-dimensional solids}.
\newblock PhD thesis, Massachusetts Institute of Technology, 1963.

\bibitem{silver2017mastering}
D.~Silver, J.~Schrittwieser, K.~Simonyan, I.~Antonoglou, A.~Huang, A.~Guez, T.~Hubert, L.~Baker, M.~Lai, A.~Bolton, et~al.
\newblock Mastering the game of go without human knowledge.
\newblock {\em Nature}, 550(7676):354--359, 2017.

\bibitem{song2017suncg}
S.~Song, F.~Yu, A.~Zeng, A.~X. Chang, M.~Savva, and T.~Funkhouser.
\newblock Semantic scene completion from a single depth image.
\newblock In {\em 2017 IEEE Conference on Computer Vision and Pattern Recognition (CVPR)}, pages 190--198, 2017.

\bibitem{statista2023vr}
{S}tatista.
\newblock "{N}umber of {V}irtual {R}eality ({VR}) and {A}ugmented {R}eality ({AR}) {U}sers in the {U}nited {S}tates {F}rom 2017 to 2023.," https://www.statista.com/statistics/1017008/united-states-vr-ar-users/, (accessed {J}an. 30, 2024).

\bibitem{sun20233dgpt}
C.~Sun, J.~Han, W.~Deng, X.~Wang, Z.~Qin, and S.~Gould.
\newblock 3d-gpt: Procedural 3d modeling with large language models, 2023.

\bibitem{tang2023dreamgaussian}
J.~Tang, J.~Ren, H.~Zhou, Z.~Liu, and G.~Zeng.
\newblock Dreamgaussian: Generative gaussian splatting for efficient 3d content creation.
\newblock {\em arXiv preprint arXiv:2309.16653}, 2023.

\bibitem{tolstikhin2021mlp}
I.~O. Tolstikhin, N.~Houlsby, A.~Kolesnikov, L.~Beyer, X.~Zhai, T.~Unterthiner, J.~Yung, A.~Steiner, D.~Keysers, J.~Uszkoreit, et~al.
\newblock Mlp-mixer: An all-mlp architecture for vision.
\newblock {\em Advances in Neural Information Processing systems}, 34:24261--24272, 2021.

\bibitem{delatorre2024llmr}
F.~D.~L. Torre, C.~M. Fang, H.~Huang, A.~Banburski-Fahey, J.~A. Fernandez, and J.~Lanier.
\newblock {LLMR}: Real-time prompting of interactive worlds using large language models.
\newblock {\em arXiv preprint arXiv:2309.12276}, 2024.

\bibitem{vrchat2023}
{VRC}hat {I}nc.
\newblock "{VRC}hat," https://hello.vrchat.com/ (accessed {J}an. 29, 2024).

\bibitem{Wang2018indoor}
K.~Wang, M.~Savva, A.~X. Chang, and D.~Ritchie.
\newblock Deep convolutional priors for indoor scene synthesis.
\newblock {\em ACM Transactions on Graphics}, 37(4), jul 2018.

\bibitem{xu2023dream3d}
J.~Xu, X.~Wang, W.~Cheng, Y.-P. Cao, Y.~Shan, X.~Qie, and S.~Gao.
\newblock Dream3d: Zero-shot text-to-3d synthesis using 3d shape prior and text-to-image diffusion models.
\newblock In {\em Proceedings of the IEEE/CVF Conference on Computer Vision and Pattern Recognition (CVPR)}, pages 20908--20918, 2023.

\bibitem{yakimovsky1973semantics}
Y.~Yakimovsky and J.~A. Feldman.
\newblock A semantics-based decision theory region analyser.
\newblock In {\em IJCAI}, volume~73, pages 580--588, 1973.

\bibitem{resonite2023}
{Yellow Dog Man Studios}.
\newblock "{R}esonite,", https://store.steampowered.com/app/2519830/resonite/ (accessed {J}an. 29, 2024).

\bibitem{yin2024text2vrscene}
Z.~Yin, Y.~Wang, T.~Papatheodorou, and P.~Hui.
\newblock Text2vrscene: Exploring the framework of automated text-driven generation system for vr experience.
\newblock In {\em Proceedings of IEEE International Conference of Virtual Reality 2024}, page preprint, 2024.

\bibitem{Yu2011makeithome}
L.-F. Yu, S.-K. Yeung, C.-K. Tang, D.~Terzopoulos, T.~F. Chan, and S.~J. Osher.
\newblock Make it home: automatic optimization of furniture arrangement.
\newblock {\em ACM Transactions on Graphics}, 30(4), jul 2011.

\bibitem{yun2019cutmix}
S.~Yun, D.~Han, S.~J. Oh, S.~Chun, J.~Choe, and Y.~Yoo.
\newblock Cutmix: Regularization strategy to train strong classifiers with localizable features.
\newblock In {\em Proceedings of the IEEE/CVF international conference on computer vision (CVPR)}, pages 6023--6032, 2019.

\bibitem{zhang2017mixup}
H.~Zhang, M.~Cisse, Y.~N. Dauphin, and D.~Lopez-Paz.
\newblock mixup: Beyond empirical risk minimization.
\newblock {\em arXiv preprint arXiv:1710.09412}, 2017.

\bibitem{zhong2020random}
Z.~Zhong, L.~Zheng, G.~Kang, S.~Li, and Y.~Yang.
\newblock Random erasing data augmentation.
\newblock In {\em Proceedings of the conference on artificial intelligence (AAAI)}, volume~34, pages 13001--13008, 2020.

\end{thebibliography}
